\documentclass{article}

\usepackage{graphicx} % For including images
\usepackage[utf8]{inputenc} % For handling special characters
\usepackage{authblk}
\usepackage[english]{babel}
\usepackage{url}

\usepackage[hidelinks]{hyperref}
\usepackage{amsfonts} 
\usepackage{amsmath}
\usepackage{amsthm}
\newtheorem{rmk}{Remark}
\newtheorem{lemma}{Lemma}

\newtheorem*{pf}{Proof}

\begin{document}
%\let\WriteBookmarks\relax
%\def\floatpagepagefraction{1}
%\def\textpagefraction{.001}

%\shorttitle{Regularized log-linear cost model}    

%\shortauthors{}  

%\title [mode = title]{Regularizing Log-Linear Cost Models for Inpatient Stays by Merging ICD-10 Codes}  
%\title{Regularizing Log-Linear Cost Models for Inpatient Stays by Merging ICD-10 Codes}
\title{A Log-Linear Analytics Approach to Cost Model Regularization for Inpatient Stays through Diagnostic Code Merging}
\author{Chi-Ken Lu\footnote{Corresponding author: CL1178@RUTGERS.EDU}}
\author{David Alonge}
\author{Nicole Richardson}
\author{Bruno Richard}
%\orcid{0000-0003-4297-2492}
\affil{Department of Mathematics and Computer Science, Rutgers University, Newark, New Jersey, USA}
%\author[1]{Alison Carefully}
%\author[2]{Ivor Question}
%\affil[1]{Department of Mathematics, University X}
%\affil{$^*$Author to whom any correspondence should be addressed.}

%\email{CL1178@rutgers.edu}
\maketitle
%\keywords{high-dimensional regression, coefficient consistency, implicit regularization, MedPAR, variables grouping, Hessian matrix}
\begin{abstract}
%Here goes the abstract \nocite{*}%% Remove this line from your manuscript.
Cost models in healthcare research must balance interpretability, accuracy, and parameter consistency; yet interpretable models often struggle to achieve accuracy and consistency. Ordinary least squares (OLS) models for high-dimensional regression can be accurate. Still, these models will fail to produce stable regression coefficients across subsamples when using highly granular ICD-10 diagnostic codes as predictors. This instability arises because many ICD-10 codes are infrequent in healthcare datasets. While regularization methods such as Ridge can address this issue, they risk discarding important predictors. Here, we demonstrate that reducing the granularity of ICD-10 codes is an effective regularization strategy within OLS while preserving the representation of all diagnostic code categories. 
%By truncating ICD-10 codes from seven characters (e.g., T67.0XXA, T67.0XXD) to six (e.g., T67.0XX) or fewer, we reduce the dimensionality of the regression problem while maintaining model interpretability and consistency. 
By truncating ICD-10 codes from seven characters to six or fewer, we reduce the dimensionality of the regression problem while maintaining model interpretability and consistency.
Mathematically, the merging of predictors in OLS leads to an increased trace of the Hessian matrix, reducing the coefficient estimation variance. Our findings shed light on why broader diagnostic groupings like DRGs and HCC codes are favored over highly granular ICD-10 codes in real-world risk adjustment and cost models.
\end{abstract}

% Use if a graphical abstract is present

%\item 
%\item 
%\item 
%\item A high-dimensional log-linear cost model using only binary ICD-10 code indicators as predictors can explain approximately 40\% of the variance in inpatient stay costs.
%\item The binary design matrix $X$ yields a Hessian matrix equivalent to the occurrence and co-occurrence matrix of ICD-10 codes.
%\item We introduce a metric to quantify the inconsistency of regression coefficients produced by unregularized linear models.
%\item Reducing the granularity of ICD-10 codes acts as a form of implicit regularization: grouping codes by truncating their character length -- without the clinical knowledge required for DRGs or HCC codes-- improves the stability of coefficient estimates. 

% Keywords
%\begin{keywords}
%high-dimensional regression\sep coefficient consistency\sep implicit regularization\sep MedPAR\sep variables grouping\sep Hessian matrix
%\end{keywords}

%\maketitle

% Main text
%\section{}\label{}

\section{Introduction}

Accurate and interpretable cost models are essential in healthcare research as they provide critical tools for estimating, analyzing, and understanding healthcare spending patterns~\cite{Duan01041983,griswold2004analyzing,Layton2017,fernandez2019estimating}. 
Linear and log-linear regression methods are a particularly popular approach to modeling healthcare data as they are relatively simple to implement and highly interpretable~\cite{James2013ISLR,kan2019exploring,irvin2020incorporating}. 
Log-linear models can easily leverage individual-level features that are prevalent in healthcare datasets (e.g., demographic and diagnostic information) to predict outcomes such as the cost of care for acute and chronic conditions, and account for the characteristics of cost distributions~\cite{Reed:2001aa,THONGPETH2021100769,sandra2023prediction,RAO2024100351}.
A prominent example is the Center for Medicare and Medicaid Services (CMS), which uses regression-based models in its risk adjustment methodologies for healthcare payment systems~\cite{Evans2011Evalutaion,kautter2014hhs}. 
These models produce regression coefficients that form the basis for risk scores to determine reimbursements for enrollees with diverse demographics and health statuses, which makes their accuracy crucial for both the patient and the provider. 

%As in many domains, healthcare data today is high-dimensional as it involves a large set of patient observations ($n$) across many features ($p$), including patient demographics~\cite{Ash2017}, provider details, financial information, and diagnostic codes. 
%This high dimensionality can hinder a model’s ability to learn stable, robust representations—an issue commonly addressed through regularization techniques as well as dimension reduction. 
%For instance, Ridge regression introduces a penalty term that shrinks coefficients to reduce model variance, mitigating overfitting at the expense of added bias~\cite{hoerl1970ridge,hastie2017elements}. 
%In some deep learning, pooling layers reduce feature dimensionality to facilitate efficient and meaningful representation learning~\cite{Goodfellow-et-al-2016}.
%While the nonlinear operations enhance expressivity of machine learning models, the interpretability, i.e. how the variables contribute to predicting the outcome, is compromised.    
%In this context, predicting the cost of inpatient stays, which account for a significant portion of total healthcare expenditures in the United States~\cite{Pfuntner2006},  is especially challenging and important as a great deal of diagnostic variables are attached to the stays. Most commonly, International Classification of Diseases, Tenth Revision, Clinical Modification (ICD-10-CM)~\cite{CMS,Omalley2005} is adopted to code the diagnoses. As such, modeling these costs with high accuracy and consistency is a shared concern of machine learning and healthcare communities.

Like many other domains, healthcare data is inherently high-dimensional, consisting of a large number of patient observations ($n$) measured across numerous features ($p$), such as demographics~\cite{Ash2017}, provider information, financial records, and diagnostic codes. This high dimensionality poses challenges for learning stable and robust representations, often necessitating the use of regularization or dimensionality reduction methods. For example, Ridge regression introduces a coefficient-shrinking penalty that reduces variance and mitigates overfitting, albeit at the cost of additional bias~\cite{hoerl1970ridge,hastie2017elements}. In deep learning, pooling layers similarly reduce feature dimensionality, enabling more efficient and meaningful representation learning~\cite{Goodfellow-et-al-2016}. However, while such nonlinear operations enhance expressivity, they often diminish interpretability, obscuring the role of individual variables in outcome prediction. These challenges are particularly pronounced in estimating the cost of inpatient stays, which represent a substantial share of U.S. healthcare expenditures~\cite{Pfuntner2006}. Because each stay is linked to numerous diagnostic variables, typically encoded using the International Classification of Diseases, Tenth Revision, Clinical Modification (ICD-10-CM)~\cite{CMS,Omalley2005}, achieving accurate and consistent cost prediction remains a critical problem at the intersection of machine learning and healthcare.

\begin{figure}[ht]
  \centering
  \includegraphics[width=.8\textwidth]{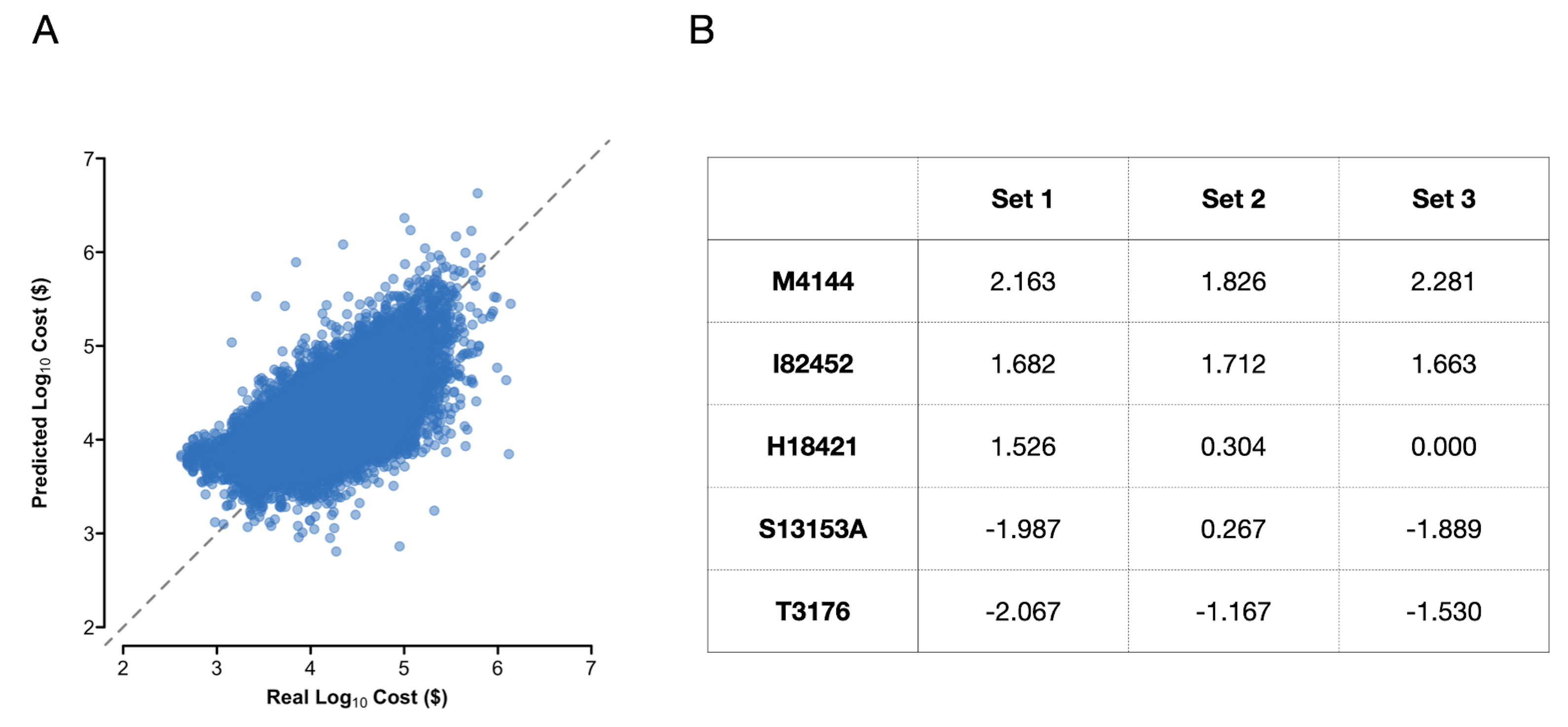}
  \caption{Panel A: the predictive log costs from OLS models against their true values. The OLS models include explainable variables: indicators of ICD-10 diagnostic codes and demographics (age, sex, and race). Panel B: the inconsistency among a few regression coefficients from OLS fitting to different training data. In Set 3, the corresponding training samples do not contain H18.421 so that the fitted coefficient is zero.}
\label{ProblemStatement}
\end{figure}

This study analyzes inpatient cost data from the New York Downstate subset of the Medicare Provider Analysis and Review (MedPAR) dataset. 
Each inpatient stay has at most 25 assigned diagnosis codes. 
The outcome variable $y$ is the log-transformed cost of the stay, and the predictors ${\bf x}$ are binary variables indicating the presence of specific ICD-10 codes.\footnote{Including indicators for age, sex, and race did not significantly alter the accuracy. Thus, we excluded the demographic variables to focus on the effects attributable to ICD-10 codes.} 
We apply ordinary least squares (OLS) regression to fit the model across randomized training subsets ($n_{\text{tr}} \approx 400{,}000$, $p \approx 20{,}000$), evaluating on a holdout test set ($n_{\text{te}} \approx 100{,}000$). The average training $R^2$ is approximately 0.45, with a test $R^2$ near 0.41. Panel A in Fig.~\ref{ProblemStatement} displays the predicted log cost for test data against its true value.
Despite the predictive scores being better than those reported for HHS-HCC risk adjustment models \cite{kautter2014hhs}, the OLS regression coefficients are highly inconsistent across subsamples. Panel B in Fig.~\ref{ProblemStatement} lists a few regression coefficients from using different training data.
This instability makes it difficult to use these coefficients for developing a reliable ICD-10-based risk score. 

\begin{figure}[ht]
  \centering
  \includegraphics[width=.8\textwidth]{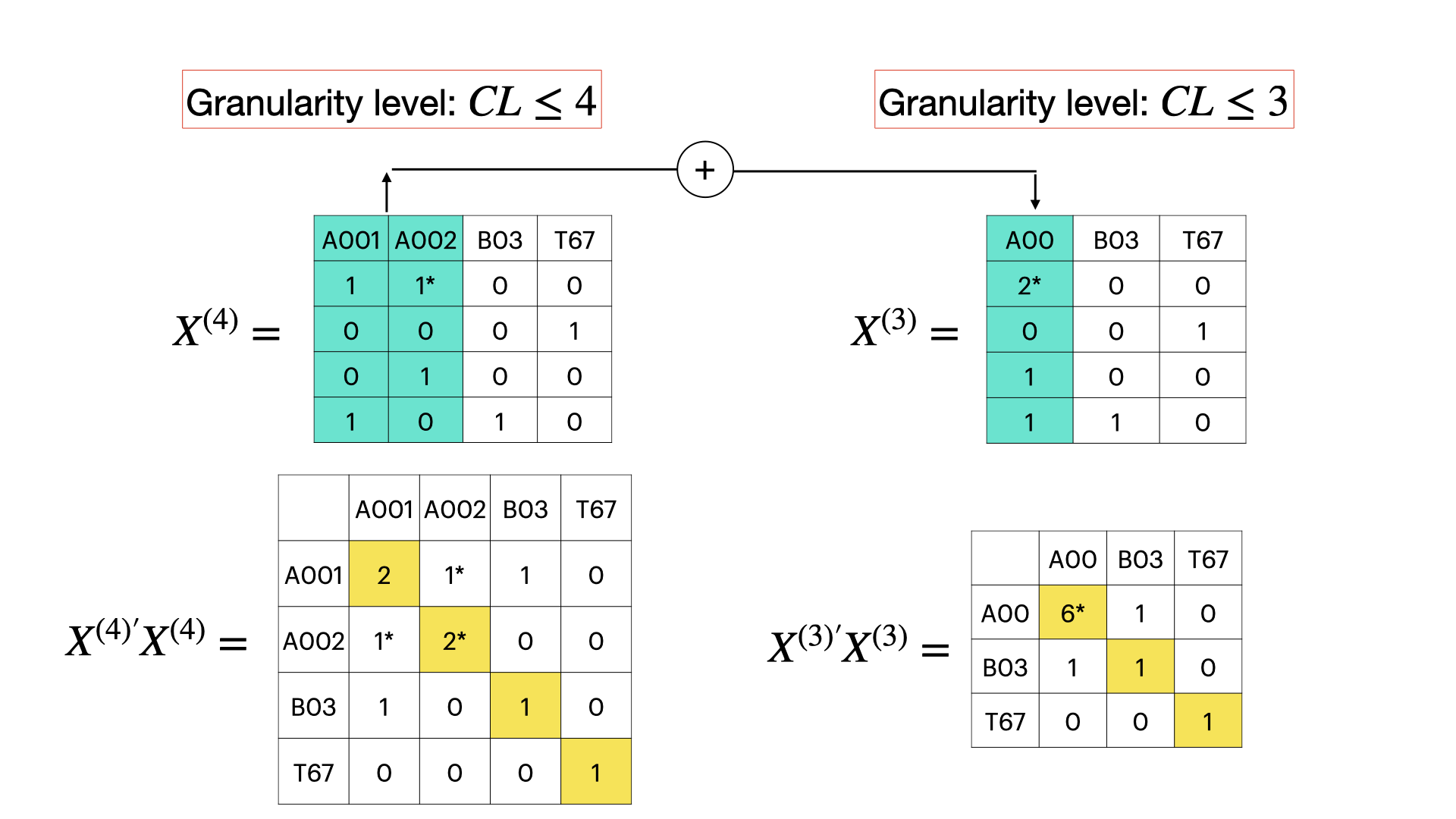}
  \caption{A toy example for illustration of reducing code granularity by merging similar codes. The design matrix $X^{(4)}$ on the left records the diagnoses using codes with $CL\leq4$. Then the granularity is lowered by truncating codes with four characters to 3. The predictors for A001 and A002 are added to form the new predictor A00 in the matrix $X^{(3)}$ on the right. The corresponding Hessian matrices are displayed at the bottom. Lowering granularity (left to right) increases the trace due to the co-occurrence of merged codes in the same stay. If the marked 1 in $X^{(4)}$ is set to 0, then the rest of the marked numbers change and the traces on both sides become identical.}
\label{GranularityReduction}
\end{figure}

The root of this inconsistency lies in the high variance of the coefficient estimates \cite{dicker2014variance}, which is governed by the diagonal entries of the inverse Hessian matrix \cite{hastie2017elements};
%\footnote{The demographic variables are not included for the purpose of simplicity as they are not relevant to the model accuracy.}, this Hessian matrix corresponds to the co-occurrence structure of ICD-10 codes in the data.
when codes appear fewer than 10 times out of 500,000 observations (i.e., extremely infrequent), the associated coefficient estimates become highly variable.
In this study, we demonstrate how Spearman's rank correlation between pairs of coefficient vectors derived from different data subsamples can be used to measure the inconsistency of regression coefficients estimated across different data subsets.
As expected, we show how using DRGs and HCCs in addition to Ridge regularization methods improves consistency of the regression coefficient estimates. 
Importantly, we propose a method of implicit regularization via ICD-10 code merging (see Fig~\ref{GranularityReduction}). 
Code truncations methods have been used to account for sparseness in and assist with dimentionality reduction in disease prediction~\cite{Shiban2021,mirtchouk2021hierarchical,qiao2022}. 
Here, we demonstrate the value of code truncation in healthcare cost prediction models, which normally approach regularization with Ridge regression or group disagnostic codes~\cite{Duan01041983,THONGPETH2021100769,technologies10060122}.
Unlike these methods, our approach reduces the dimensionality of the data while preserving the general semantic structure of diagnostic codes.  

%\textbf{this is methods}. 
We define code granularity by a maximum character length $l$, (CL) $\leq l$. 
Merging codes under a reduced granularity decreases the number of predictors ($p^{(l)} < p^{(l+1)}$) and results in a transformed design matrix with increased trace in the associated Hessian: 

\begin{equation}
	\text{tr}(X^{(l+1)\prime} X^{(l+1)}) \leq \text{tr}(X^{(l)\prime} X^{(l)}),
\end{equation}
with equality only if the merged codes have zero co-occurrence. 
This trace increase is linked to improved stability in coefficient estimation. 
Figure~\ref{GranularityReduction} illustrates this merging process and highlights a case where the equality condition holds.
By integrating these ideas, we present a framework that enhances the reliability of cost estimation models using ICD-10 codes, with implications for both healthcare policy modeling and robust machine learning in high-dimensional settings.

%\textbf{Is this necessary?}
%The generalizable insights about machine learning in the context of healthcare are summarized in the following:

%\begin{itemize}
%	\item A high-dimensional log-linear cost model using only binary ICD-10 code indicators as predictors can explain approximately 40\% of the variance in inpatient stay costs.
%	\item The binary design matrix $X$ yields a Hessian matrix equivalent to the occurrence and co-occurrence matrix of ICD-10 codes.
%	\item We introduce a metric to quantify the inconsistency of regression coefficients produced by unregularized linear models.
%	\item Reducing the granularity of ICD-10 codes acts as a form of implicit regularization: grouping codes by truncating their character length -- without relying on clinical knowledge -- improves the stability of coefficient estimates.    
%\end{itemize}

Our work develops and evaluates a high-dimensional log-linear model that predicts inpatient stay costs using binary ICD-10 code indicators, achieving approximately 40\% variance explained. We further examine the structural properties of the design matrix, introduce a new metric to quantify coefficient instability in unregularized linear models, and demonstrate that reducing ICD-10 code granularity acts as implicit regularization—enhancing model robustness. Together, these contributions improve predictive performance and model reliability in healthcare cost modeling.

%The paper has the following structures. In Sec.~\ref{literature}, relevant previous works on modeling inpatient cost using diagnosis codes are reviewed. Along with a descriptive analysis of the data subset, the subtle structure of design and Hessian matrices associated with ICD-10 codes, the variable groupings by code truncation as an implicit regularization to OLS, the variance of estimating OLS coefficients and its relation to the Hessian matrix, a metric measuring the coefficient consistency, and characteristics of Hessian matrix for the dats subset are presented in Sec.~\ref{method}. Coefficient consistency, which is an equally important factor as accuracy for cost models, is displayed in Sec.~\ref{results} as the granularity level of ICD-10 codes vary. Sec.\ref{discussions} provides insights into the results of our OLS modeling using the code truncation and limitations. A final conclusion including future questions can be seen in Sec.\ref{conclusion}. 

The paper is organized as follows. Section~\ref{literature} reviews prior studies on modeling inpatient costs using diagnosis codes. Section~\ref{method} presents a descriptive analysis of the data subset, examines the structure of the design and Hessian matrices associated with ICD-10 codes, and discusses variable groupings by code truncation as an implicit regularization method for OLS (see proofs in Appendix). It also explores the variance of OLS coefficient estimates in relation to the Hessian matrix, introduces a metric for measuring coefficient consistency, and analyzes Hessian matrix characteristics for the dataset. Section~\ref{results} reports coefficient consistency—an equally important factor as predictive accuracy in cost models—across varying levels of ICD-10 code granularity. Section~\ref{discussions} offers insights into the OLS modeling results under code truncation and discusses limitations. Finally, Section~\ref{conclusion} summarizes the findings and outlines future research directions.

\section{Related Work}\label{literature}

Historically, linear regression has been a cornerstone of modeling healthcare cost and other outcomes due to its simplicity and interpretability~\cite{Duan01041983,THONGPETH2021100769,technologies10060122}. 
Models for diagnosis-based healthcare cost may often include a large number of predictors, and dimension reduction methods are commonly employed to account for the inherent sparsity that accompanies these data.
\cite{THONGPETH2021100769} analyzed cost data from Thailand with features including demographic variables (age and gender), discharge status, and ICD-10 codes grouped into 18 diagnostic categories. 
Grouped diagnostic variables, such as those from Diagnosis-Related Groups (DRGs)~\cite{Fetter1986,Lynk2001} and Hierarchical Condition Categories (HCCs)~\cite{wagner2016,Kim2023}, are frequently used alongside demographic and hospital-specific features to account for clinical complexity. 
These groupings are especially prominent in risk adjustment frameworks like those implemented by the Centers for Medicare \& Medicaid Services (CMS)~\cite{kautter2014hhs}.

%Additionally, the group Lasso, introduced by~\cite{yuan2006model}, provides a principled approach to variable selection when predictors are organized into {\it predefined} groups. 

Ordinary Least Squares (OLS) regression is often ill-suited for modeling healthcare costs, which are typically right-skewed and heteroscedastic~\cite{THONGPETH2021100769,sandra2023prediction}. 
While log-transforming the outcome variable can partially address these issues~\cite{MANNING2001461}, high-dimensional feature spaces and nonlinear interactions among predictors can still degrade model performance.

Ridge~\cite{hoerl1970ridge} and Lasso~\cite{Tibshirani1996,bakin1999adaptive,yuan2006model} regularization introduce $L_2$ and $L_1$ penalties to reduce overfitting and multicollinearity, with Lasso further inducing sparsity as a form of variable selection. However, linear models often fail to capture the nonlinear relationships common in healthcare data; although interaction terms can approximate comorbidities, they remain limited in flexibility. To improve predictive accuracy, machine learning (ML) methods have gained increasing attention~\cite{kan2019exploring,irvin2020incorporating,Shiban2021,qiao2022,Taloba2022,Langenberger2023,sandra2023prediction}. 
\cite{RAO2024100351} studied the cost models and analyzed the anonymous patient data from the New York State Statewide Planning and Research Cooperative System, identified APR DRG as the best feature from random forest (RF) regressor.
\cite{yang2018} explored the insurance claim data from Texas Medicaid Program and used OLS, Lasso, gradient boosting (GB), and recurrent neural network to predict prospective medical expenditures. The variables include grouped diagnostic codes, medication, and demographic informations.
\cite{RAHMAN2025100411} used linear and tree-based models for predicting clinical service utilization by analyzing medical records from the Louisiana Department of Health.
\cite{THONGPETH2021100769} reported that RF outperformed Ridge regression in predicting the associated cost.

Tree-based models, such as decision trees, random forests, and gradient boosting, partition the input space hierarchically. RF mitigate overfitting by averaging predictions across bootstrapped ensembles, while GB refines predictions sequentially, often achieving superior performance at the cost of interpretability. Deep learning models, in turn, serve as universal function approximators capable of capturing complex nonlinear patterns. A multi-layer perceptron (MLP), for instance, employs stacked layers of interconnected nodes that iteratively optimize their weights to learn abstract data representations~\cite{hastie2017elements}. 

While machine learning models are designed to capture complex relationships between predictors and outcomes, their development often prioritizes accuracy over interpretability and stability. In the context of healthcare data, however, interpretability—such as understanding how diagnostic variables influence medical expenditures—is as critical as predictive performance. Black-box algorithms hinder providers, policymakers, and patients from identifying the drivers of healthcare costs or leveraging model insights to address systemic inequities. Beyond their interpretability, linear models have the advantage of well-established variance analysis~\cite{dicker2014variance,mel2021theory,hastie2022surprises}, which is essential for studying coefficient consistency in diagnosis-based inpatient cost modeling. By contrast, conducting variance analysis for more complex machine learning models remains largely intractable.

\section{Methods}\label{method}

%This section begins with a description of the inpatient stay data extracted from the Downstate New York subset of the Medicare Provider Analysis and Review (MedPAR) dataset. 
%We summarize key characteristics of the data, including the distribution of stay costs and the number of ICD-10 diagnosis codes assigned per stay.
%\textbf{This is introduction material}
%\textit{
The primary modeling approach is a diagnosis-based log-linear regression, where the log-transformed inpatient cost is regressed on high-dimensional binary variables, indicating the presence of specific ICD-10 codes. 
A central challenge in this setting is the inconsistency of coefficient estimates obtained from Ordinary Least Squares (OLS) when the model is repeatedly fit on random subsamples of the data. 
This inconsistency stems from the high variance in coefficient estimation, which is closely linked to the eigenvalues of the Hessian matrix derived from the squared loss function.

%\textit{
To address this issue, we examine two strategies for improving coefficient stability. 
First, we analyze how Ridge regression mitigates inconsistency by adding a penalty term to the Hessian, increasing its eigenvalues and reducing variance. 
Second, we introduce a novel approach: merging similar ICD-10 codes to reduce model dimensionality. 
We show that this merging process also leads to an increase in the trace of the Hessian (i.e., the sum of its eigenvalues), resulting in improved consistency across model fits.
%}

\subsection{Data}
\begin{figure}[h]
  \centering
\includegraphics[width=1\textwidth]{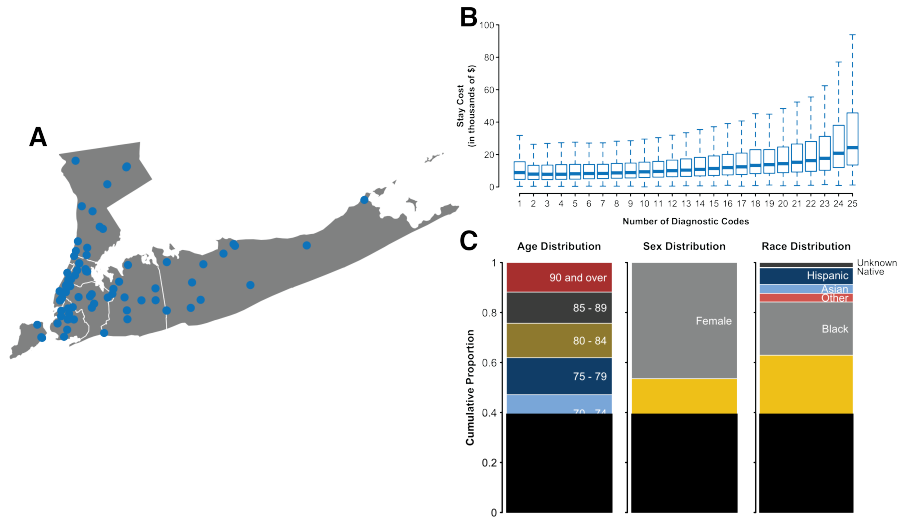}
 \caption{\textbf{A}. Map of the hospitals that make up the downstate New York subset we use in our analysis. 
 %\textbf{B}. Distributions of size (number of beds) and total costs for hospitals in the greater USA, Upstate New York, and Downstate New York in the FY2018 subset. 
 \textbf{B}. Association between the average cost of a stay and the number of diagnostic codes attached to the stay for hospitals in Downstate New York. The more diagnostic codes are attached to a stay, the more expensive the stay.
 \textbf{C}. Distributions over the Age, Sex, and Race variables in the MedPAR subset.}
  \label{HospitalDemographics}
\end{figure}

The Medicare Provider Analysis and Review (MedPAR) Limited Data Set contains detailed information on all Medicare and Medicaid beneficiaries who utilize hospital inpatient services in the United States. 
Each record in MedPAR corresponds to an individual inpatient stay, which aggregates claims submitted throughout a continuous hospitalization from the date of admission to the date of discharge.
For each stay, MedPAR holds variables that describe the demographics of patients, the healthcare provider, financial information (e.g., total charges), diagnostic and procedural codes, and time-related fields such as admission and discharge dates.
%\textbf{
In this study, we estimate the cost of patient stays in a subset of the MedPAR FY2018 file. 
To manage the scale of data while preserving sufficient sample size and clinical diversity, we subset the data geographically to include only providers located in Downstate New York, which comprises Westchester, Bronx, New York (Manhattan), Queens, Kings (Brooklyn), Richmond (Staten Island), Nassau, and Suffolk counties (see Figure~\ref{HospitalDemographics}-A).
This subset reduced the size of the MedPAR data to approximately 3\% of its original size per year.
%}% ($N_{2018} = 495,617$, $N_{2019} = 565,787$, $N_{2020} = 494,427$).

While the Downstate New York subsets of the MedPAR data are broadly comparable to the full national dataset, some differences emerge in cost and demographic compositions.
Hospitals in Downstate New York are, on average, larger than those in both Upstate New York and the broader United States and report higher average costs per inpatient stay. 
The downstate New York region includes fewer White patients and higher proportions of Black, Asian, and Hispanic patients compared to the national MedPAR dataset, while there are no meaningful differences in the patient age and sex distributions.

\subsection{Ordinary Least Square and Binary Variables for ICD Codes}\label{OLS}

Let $y = \log_{10}(\text{cost})$ represent the logarithm (base 10) of the inpatient stay cost. This quantity is modeled as a linear function of $p$ binary predictors ${x_1, x_2, \dots, x_p}$, which indicate the presence of corresponding ICD-10 diagnosis codes ${\alpha_1, \alpha_2, \dots, \alpha_p}$. Up to a zero-mean noise term $\epsilon$, the model takes the form

\begin{equation}
y = \beta_0 + \sum_{i=1}^p \beta_i x_i + \epsilon, \label{log-linear-model}
\end{equation}
where $\beta = {\beta_0, \beta_1, \dots, \beta_p}$ are the regression coefficients to be estimated.
The coefficients are obtained by minimizing the ordinary least squares (OLS) objective function:
\begin{equation}
\mathcal{L}_{\text{OLS}} = \sum_{i=1}^n (y_i - \hat{y}_i)^2, \label{squared_loss}
\end{equation}
where $n$ is the number of observations. 
The closed-form OLS solution is given by
\begin{equation}
\hat{\beta} = (\tilde{X}^\prime\tilde{X})^{-1} \tilde{X}^\prime y, \label{OLS_weights}
\end{equation}
where $\tilde{X}$ is the design matrix augmented with an intercept column
\begin{equation}
\tilde{X} = \begin{bmatrix}
X_{(n \times p)} & \mathbf{1}_{(n \times 1)}
\end{bmatrix}, \label{matrix_form}
\end{equation}
The corresponding $(p+1) \times (p+1)$ Hessian matrix of the squared loss objective is
\begin{equation}
\tilde{X}’\tilde{X} =
\begin{bmatrix}
X^\prime X & X^\prime \mathbf{1} \\
\mathbf{1}^\prime X & n
\end{bmatrix}. \label{block_form}
\end{equation}

The variance-covariance matrix of the estimated coefficients is %$\hat{\beta}$ is
%\begin{equation}
%$\operatorname{Var}(\hat{\beta}) = 
$\hat{\sigma}^2 (\tilde{X}^\prime \tilde{X})^{-1}$%\:,
%\end{equation}
where $\hat{\sigma}^2$ is the estimated variance of the noise term $\epsilon$. The variance of an individual coefficient $\hat{\beta}_i$ is therefore
\begin{equation}
\operatorname{Var}(\hat{\beta}_i) = \hat{\sigma}^2 v_i\:,
\end{equation}
with $v_i$ being the $i$-th diagonal element of $(\tilde{X}^\prime \tilde{X})^{-1}$ \cite{hastie2017elements}. This term directly reflects the sensitivity of coefficient estimates to the structure and sparsity of the design matrix—a key factor in the inconsistency of regression coefficients in high-dimensional OLS models.

Statistically, a large value of $v_i$ indicates that the estimate $\hat{\beta}_i$ will vary significantly when different subsets of the data are used to fit the model. 
Unfortunately, the diagonal entries of an inverse matrix are highly nonlinear functions of the entries in the original matrix, which makes it challenging to control the variance directly.
To mitigate this, one can explicitly add a penalty term to the loss function—such as in Ridge regression—which increases the eigenvalues of the Hessian and thereby reduces all variances $v_i$.
We present the following remark, which provides an important observation regarding the total variance of the coefficients and its relationship to the eigenstructure of the design matrix.

\begin{rmk} Sum of coefficient variance $S_V=v_1+v_2+\cdots+v_p+v_0$, sum of eigenvalues of inverted Hessian $S_I=s_1^{-1}+s_2^{-1}+\cdots s_p^{-1}+s_0^{-1}$, and trace of Hessian satisfy the following relations,
\begin{equation}
	S_V=S_I>\frac{1}{tr({\tilde X}'{\tilde X})}\:.\label{lower_bound}
\end{equation}
\label{remark:1}
\begin{pf} %of Remark~\ref{remark:1}. 
Let the set $\{s_1,s_2,\cdots,s_p,s_0\}$ denote the eigenvalues of the Hessian ${\tilde X}'{\tilde X}$. It follows that the inverse, $({\tilde X}'{\tilde X})^{-1}$, has the set of eigenvalues $\{s_1^{-1},s_2^{-1},\cdots,s_p^{-1},s_0^{-1}\}$. 
Using the fact that the sum of eigenvalues of a square matrix is identical to the sum of its diagonal entries, then the equality follows as the square matrix is the inverse of Hessian. Since the Hessian is positive definite, all eigenvalues are positive. Since $f(s)=1/s$ is convex for $s>0$, we can use Jensen's inequality to show $S_I=\mathbf E[1/s]>1/\mathbf E[s]$. Note that the sums have been replaced with an expectation. Finally, $s_1+s_2+\cdots+s_p+s_0=\mathbf E[s]=tr({\tilde X}'{\tilde X})$, which completes the proof.
\end{pf}
\end{rmk}

This highlights that both the eigenvalues and the diagonal elements of the Hessian matrix play a critical role in determining the stability of coefficient estimates and the model's predictive performance. Therefore, understanding what structural properties of the design matrix contribute to large variance $v$ values is essential.

We now examine the structure of the matrix $X^{\prime}X$, which is the main component of the block matrix in Eq. \eqref{block_form}. Define the frequency $F_j$ of ICD code $j$ as the column sum $F_j := \sum_i X_{ij}$, and define $D_i$ as the number of diagnoses recorded for stay $i$, i.e., the row sum $D_i := \sum_j X_{ij}$. Due to MedPAR formatting, $D_i \leq 25$, which results in a highly sparse matrix $X$.

\begin{rmk}
The ${p\times p}$ matrix $X'X$ is, in fact, the co-occurrence matrix for the ICD codes. That is the $j$th diagonal entry $[X'X]_{jj}=F_j$, the total frequency of the ICD code $j$. The off-diagonal entries $[X'X]_{jk}$ represent the co-occurrence of the ICD code $j$ and code $k$. Consequently, the sum of eigenvalues $\sum_{i=1}^ps_i=\sum_{i=1}^pF_i$.
\label{remark:2}
\begin{pf} %of Remark~\ref{remark:2}. 
The diagonal entry $[X'X]_{jj} = \sum_i(X_{ij})^2$, the $L^2$-norm of the binary column $X_{:,j}$. As 0 and 1 are invariant under square, $[X'X]_{jj}=\sum_iX_{ij}=F_j$. Similarly, off-diagonal entry $[X'X]_{jk}=\sum_iX_{ij}X_{ik}$, a count of stays having with ICD-10 codes $j$ and $k$ simultaneously in the record. As the diagonal entries are connected with the ICD code frequencies, using the property of trace can show that the sum of ICD-10 code frequencies in the data equal the sum of eigenvalues. 
\end{pf}
\end{rmk}

For example, in the FY-2018 MedPAR subset, there are approximately 500,000 stays and over 20,000 unique ICD-10-CM codes. 
The frequencies of individual codes vary widely, ranging from 1 to over 300,000. 
The average frequency per stay per code, computed as $\text{tr}(X^{\prime}X)/(np)$, is approximately 0.00073, reflecting the sparsity of the data.

In the following sections, we describe both explicit and implicit techniques for increasing the trace of the Hessian matrix. 
These techniques effectively reduce the variance of the estimated regression coefficients, thereby improving the consistency and reliability of the log-cost model.

\subsection{Explicit Regularization and Effective Dimension}
By adding an $L2$-norm penalty $\lambda\sum\beta_i^2$ to the original objective $\mathcal L_{OLS}$,  we obtain the Ridge regression estimator 
\begin{equation}
    \hat{\beta}_{Ridge} = ({\tilde X}'{\tilde X}+\lambda I_{p+1})^{-1}{\tilde X}'{\rm y}\:,\label{Ridge_weights}
\end{equation} 
where $\lambda > 0$ is the regularization parameter, and $I_{p+1}$ is the identity matrix of size $(p+1) \times (p+1)$. 
The added penalty term shrinks the coefficients toward zero, resulting in estimates typically closer to zero than those from OLS.
The lower bound for the sum of coefficient variances in Eq.~\eqref{lower_bound} is also affected by regularization, becoming: $1/[tr({\tilde X}'{\tilde X})+(p+1)\lambda]$. 
Thus, increasing $\lambda$ decreases the total variance of the estimated coefficients $\hat{\beta}$, enhancing their stability.

%Lasso is another method of explicit regularization. A $L1$-norm of the coefficients $\lambda_L\sum|\beta_i|$ is added as penalty to the objective function in Eq.~(\ref{squared_loss}). The $L1$-norm promotes sparsity for the regression coefficients. The coefficient $\beta_i$ for code $i$ vanishes if the correlation between the ith column $\tilde X_{:,i}$ and the residual ${\rm y}-{\tilde X}_{-i}\beta_{-i}$ is weaker than the penalty strength $\lambda_L$ [see Ch.13 in~\cite{murphy2012machine}]. The notation ${\tilde X}_{-i}$ means the same matrix of $\tilde X$ but with the ith column being removed. Similarly, $\beta_{-i}$ means that the coefficient $\beta_i$ is dropped.

The expected prediction error (risk) $\mathbb{E}[(y - \mathbf{x}^{\prime}\hat{\beta})^2]$ can be decomposed into three components: the irreducible noise, squared bias, and variance \cite{hastie2017elements}. 
Regularization reduces variance at the cost of increased bias since the penalty term pulls estimates away from their true values. 
In effect, this reduces model complexity.
The effective dimension of the Ridge model, which quantifies the model’s complexity, is given by the trace of the matrix S that maps the observed responses to the predicted ones: $\hat{\mathbf{y}} = S \mathbf{y}$ \cite{hastie2017elements, maddox2020rethinking},

\begin{equation}
	\rho = tr[\tilde X({\tilde X}'\tilde X+\lambda I)^{-1}\tilde X']=\sum_{i=0}^p\frac{s_i}{s_i+\lambda}\:,\label{effD}
\end{equation} 
where $s_i$ are the eigenvalues of $\tilde{X}^\prime\tilde{X}$.

Although $\rho$ provides insight into model capacity, computing it exactly requires knowledge of all eigenvalues, which can be computationally expensive. However, the function $f(s) = \frac{s}{s + \lambda}$ is concave, allowing for a convenient upper bound.
\begin{lemma}
The effective dimension for the Ridge regression model trained on $n$ data has a upper bound $\rho_B$,
\begin{equation}
	\rho\leq\rho_B=\frac{(p+1)\bar{s}}{\bar{s}+\lambda/n}\:.\label{upperBoundDim}
\end{equation}
where $\bar{s}$ stands for the average of the eigenvalues of $\tilde{X}^{\prime}\tilde{X}/n$.

\label{lemma:1} 
\begin{pf} %of Lemma~\ref{lemma:1}. 
As the number of predictors $p\gg1$, the effective dimension can be evaluated as an integral, $\rho\approx(p+1)\int ds\frac{s}{s+\lambda}p(s)=(p+1)\mathbf E[\frac{s}{s+\lambda}]$ where $p(s)$ stands for the density of eigenvalues of Hessian matrix. Since the function $f(s)=\frac{s}{s+\lambda}$ is concave, i.e. $f''(s)<0$ for all $s>0$ provided $\lambda>0$. The above inequality directly follows Jensen's inequality.
\end{pf}
\end{lemma}
This upper bound $\rho_B$ is much easier to compute and is useful for analyzing and comparing predictive performance across different levels of regularization.

\subsection{Consistency Metric}

We repeatedly split the data into training and test sets to assess consistency, estimating regression coefficients $\beta$ each time. If the estimates are consistent, then coefficient pairs ($\beta^{(a)}$, $\beta^{(b)}$) from different splits should be relatively equal. We treat these pairs as samples from a bivariate distribution and measure their agreement using Spearman correlation, which is less sensitive to outliers. We define the consistency metric $\eta$ as the mean Spearman correlation over all distinct pairs
\begin{equation}
\eta = \frac{1}{N(N-1)}\sum_{a\neq b} r_s(\beta^{(a)},\beta^{(b)})\label{consistency_metric}
\end{equation}
Higher values of $\eta$ (close to 1) indicate strong consistency, while lower values suggest instability, with coefficient scatter resembling a circle. 

\subsection{Varying Code Granularity as Implicit Regularization}

ICD-10-CM codes have character lengths of 7, beginning with an uppercase letter (A-Z), indicating the broad disease category, followed by two numbers that indicate the disease within the broad category. This is followed by 3-4 characters. To maintain semantic meaning, the merging of ICD-10 codes can therefore have a code length (CL) of 3 to 7 (i.e., $3 \leq \text{CL} \leq 7$).
In our dataset, the set of codes $\alpha$ contains approximately $p = |\alpha| \approx 20,000$ unique ICD-10 codes (see Sec. \ref{OLS}). 
To reduce the number of predictors $p$, we propose shortening longer codes by truncating them to a fixed length $l$, merging similar codes, and reducing dimensionality.
Specifically, for any code with $\text{CL} > l$, we retain only its first $l$ characters.
In this study, we explore the impact of truncating ICD-10 codes to have $l$ between 2 and 7 characters, whereby 7 is the full ICD-10 diagnostic code. 
Reduced sets after truncation are denoted by $\alpha^{(l)}$, with $p^{(l)} = |\alpha^{(l)}|$. 
We define the truncation function
\begin{equation}
T^{(\text{CL} \leq l)}(\text{code}) = \text{code}[0:l],
\end{equation}
using Python-style string slicing. 
The granularity of the codes is controlled by $l$; decreasing $l$ reduces the number of unique codes $p^{(l)}$. 
For instance, $T^{(\text{CL} \leq 6)}(\text{T670XXA}) = T^{(\text{CL} \leq 6)}(\text{T670XXD}) = \text{T670XX}$, effectively merging both codes into a single one in $\alpha^{(6)}$. 
The progression $\alpha \rightarrow \alpha^{(6)} \rightarrow \cdots \rightarrow \alpha^{(2)}$ is analogous to hierarchical reductions used in ICD-9 tree structures \cite{mirtchouk2021hierarchical}. 
%The work in \cite{doi:10.1200/CCI.21.00186} used a similar truncation scheme to ICD-9 codes in reducing sparseness in the model predicting survival after traumatic aortic injury; \cite{Shiban2021.07.28.21261166} reduced ICD-10 codes to three characters in the model predicting hospital mortality among older patients with cancer.

Next, we describe how the predictors $x_i$ and matrix $X$ evolve under this dimensionality reduction. 
Drawing an analogy from convolutional neural networks (CNNs) \cite{Goodfellow-et-al-2016,aich2018global,sun2017learning}, where pooling layers aggregate pixels into coarser representations, we define a similar “sum-pooling” operation for variables. 
Specifically, the coarser variable $x_i^{(l)}$ is computed as a sum of finer variables
\begin{equation}
x_i^{(l)} = \sum_j Q^{(m \rightarrow l)}_{ji} x_j^{(m)},
\end{equation}
where $m > l$, and the matrix $Q^{(m \rightarrow l)} \in \mathbb{R}^{p^{(m)} \times p^{(l)}}$ has entries $Q_{ji} = 1$ if the truncated code $T^{(\text{CL} \leq l)}(\alpha_j^{(m)}) = \alpha_i^{(l)}$, and 0 otherwise. 
This operation merges columns in $X^{(m)}$ to produce the coarser matrix $X^{(l)}$, preserving the total number of diagnoses per stay (i.e., row sums remain unchanged). 
The association of cost and number of codes attached to stay (Fig.~\ref{HospitalDemographics}-{\bf C}) is retained.
Figure \ref{GranularityReduction} illustrates this process using a toy matrix with $CL \leq 4$, reduced to $CL \leq 3$ by merging columns.
The regression coefficients $\beta^{(l)}$ corresponding to $\alpha^{(l)}$ are estimated via
\begin{equation}
y = \beta^{(l)}_0 + \sum_{i=1}^{p^{(l)}} \beta^{(l)}_i x_i^{(l)} + \epsilon,
\end{equation}
with OLS still applicable. However, we now use the transformed design matrix
\begin{equation}
X^{(l)} = X^{(m)} Q^{(m \rightarrow l)},
\end{equation}
replacing the one in Eq. (\ref{matrix_form}). 
The tilded matrix ${\tilde X}^{(l)} = [X^{(l)},\mathbf{1}]$ maintains the same number of rows but has fewer columns than ${\tilde X}^{(l+1)}$, reflecting the reduced granularity.
Importantly, reducing granularity increases the trace of the Hessian matrix, which influences the variance of regression coefficients:
\begin{lemma}
The trace of the Hessian matrix increases as granularity decreases
\begin{equation}
\text{tr}(X^{(l)\prime} X^{(l)}) \geq \text{tr}(X^{(l+1)\prime} X^{(l+1)}),
\end{equation}
with equality only if codes in $\alpha^{(l+1)}$ that map to the same code in $\alpha^{(l)}$ never co-occur.
\label{lemma:2}
\begin{pf} %of Lemma~\ref{lemma:2}. 
The proof essentially bases on the observation that two non-negative columns have $({\rm x}_1+{\rm x}_2)\cdot({\rm x}_1+{\rm x}_2)\geq {\rm x}_1\cdot{\rm x}_1+{\rm x}_2\cdot{\rm x}_2$ with equality holds if ${\rm x}_1\cdot{\rm x}_2=0$. Use the trace formula, $tr(X^{(l)'}X^{(l)})=\sum_{jkm}X^{(l+1)}_{jk}X^{(l+1)}_{jm}\sum_iQ_{ki}Q_{mi}$. For ease of notation, we have dropped the superscript in $Q$. The latter sum results in the entry $(QQ')_{kl}$, representing whether the more granular code $\alpha^{(l+1)}_k$ and code $\alpha^{(l+1)}_m$ are combined into the same code in the less granular code set $\alpha^{(l)}$. Then the matrix $QQ'$ is sparse and the diagonal of it is all ones. Thus, $tr(X^{(l)'}X^{(l)})=\sum_{jkm} X^{(l+1)}_{jk}X^{(l+1)}_{jm}[I_{km}+(QQ'-I)_{km}]$ from which we conclude that $tr(X^{(l)'}X^{(l)})\geq tr(X^{(l+1)'}X^{(l+1)})$ and the equality holds when the contribution $\sum_j X^{(l+1)}_{jk}X^{(l+1)}_{jm}$ from any combined pair $(k,m)$ is zero.
\end{pf}
\end{lemma}
This is illustrated in Fig. \ref{GranularityReduction}, where reducing from $CL \leq 4$ to $CL \leq 3$ increases the trace from 6 to 8. 
If codes A001 and A002 did not co-occur in the same row, the trace would remain unchanged during reduction.

\subsection{Characteristics of Hessian matrices}
\begin{figure}[ht]
  \centering
  \includegraphics[width=0.8\textwidth]{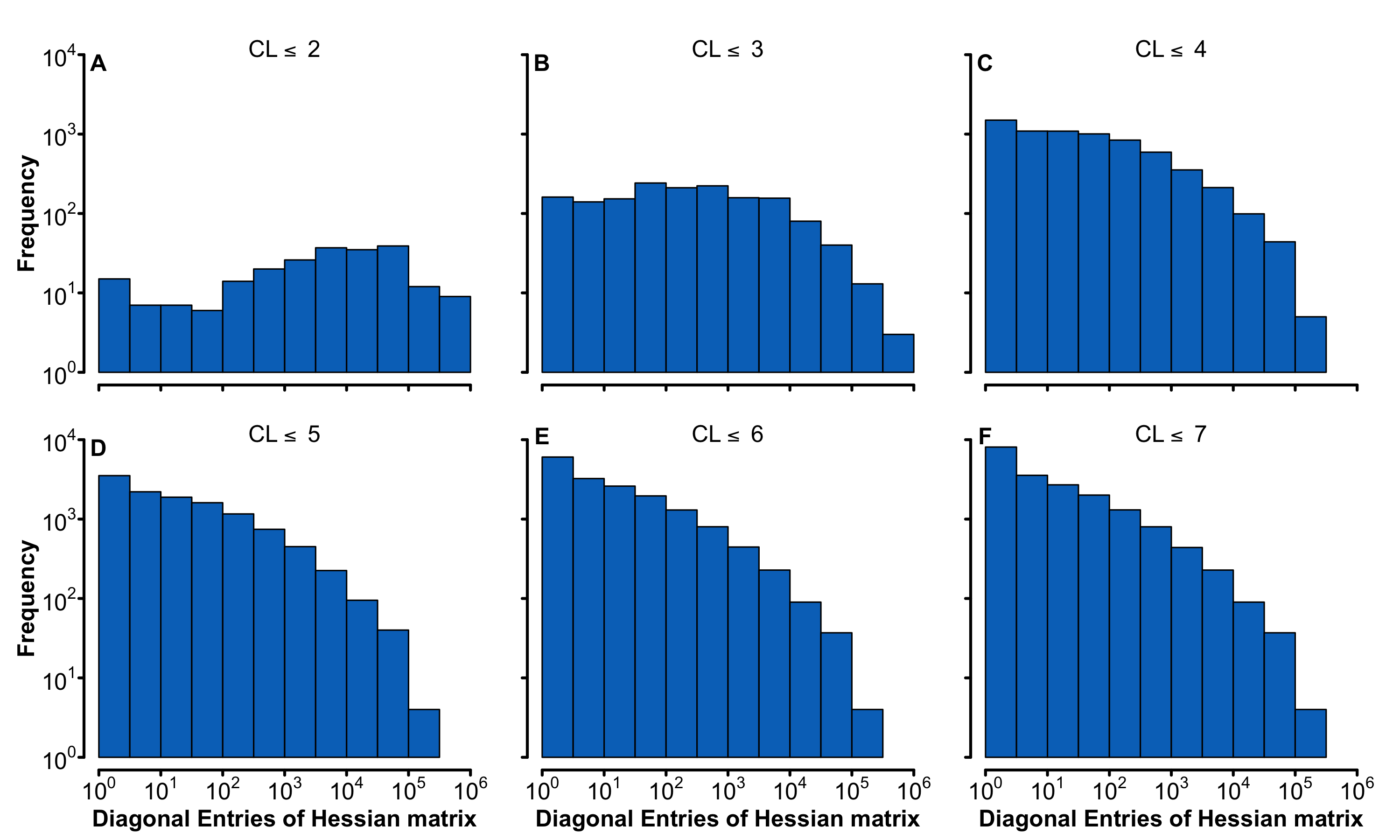}
    \caption{Evolving histograms of diagonal entries of Hessian matrix for different code granularity $CL\leq l=[2,3,4,5,6,7]$ (panel A to F). In panels D-F for higher granularity, the diagonal entries display power law distributions.}
  \label{HessianDigHistVSLetters2018}
\end{figure}

At the highest granularity level ($l$ = 7), the design matrix $X^{(7)}$ is sparse and binary. 
Efficient computation of the Hessian matrix $X^{(7)\prime}X^{(7)}$ is enabled by NumPy’s sparse matrix operations \cite{harris2020array}. 
As noted in Remark \ref{remark:2}, the diagonal entries of this Hessian matrix correspond to the frequencies of individual codes in $\alpha^{(7)}$.
The panel {\bf F} of Fig. \ref{HessianDigHistVSLetters2018} shows a log-log histogram of these diagonal entries, which follows a power-law distribution with exponent $\alpha \approx 1.93$~\cite{clauset2009power}. 
This behavior, reminiscent of word frequency distributions in natural language~\cite{clauset2009power}, suggests strong inter-code correlations~\cite{mahoney2019traditional}.

We construct design matrices $X^{(l)}$ for $l = 2, 3, \ldots, 7$ using a custom Python function. 
The corresponding log-log histograms of the Hessian diagonals are also shown in Fig. \ref{HessianDigHistVSLetters2018}. 
As granularity decreases (i.e., $l$ becomes smaller), the number of columns $p^{(l)}$ is reduced due to code merging.
While computing the full eigenvalue spectrum of the Hessian (needed for the effective dimension $\rho$ in Eq. \ref{effD} for Ridge regression) is computationally intensive, its trace -- equal to the sum of the diagonal entries -- can be efficiently computed. 
Table \ref{table:character} lists the values of $p^{(l)}$ and traces of the Hessian matrices for the FY2018 dataset. Consistent with Lemma \ref{lemma:2}, the trace increases as $l$ decreases. 
The mean eigenvalue, $\bar{s}^{(l)} := \frac{\text{tr}(X^{(l)\prime}X^{(l)})}{n p^{(l)}}$, also captures this trend,
which is included in Table \ref{table:character} and used to estimate the upper bound $\rho_B$ for the effective dimension, as discussed in Lemma \ref{lemma:1}.

\begin{table}[]
\caption{Data characteristics: the number of predictors $p^{(l)}$ and the mean eigenvalue ${\bar s}^{(l)}$ of scaled Hessian matrix $X'X/n$ with $n=495023$ observation in FY2018 subset. The ICD code granularity is specified by the character length's upper limit, $CL\leq l$.}
\begin{tabular}{|cl|cl|cl|cl|cl|cl|}
\hline
\multicolumn{2}{|c|}{CL$\leq$2}                               & \multicolumn{2}{c|}{CL$\leq$3}                               & \multicolumn{2}{c|}{CL$\leq$4}                               & \multicolumn{2}{c|}{CL$\leq$5}                               & \multicolumn{2}{c|}{CL$\leq$6}                               & \multicolumn{2}{c|}{CL$\leq$7}                               \\ \hline
\multicolumn{1}{|l|}{$p^{(2)}$} & $\bar{s}^{(2)}$             & \multicolumn{1}{l|}{$p^{(3)}$} & $\bar{s}^{(3)}$             & \multicolumn{1}{l|}{$p^{(4)}$} & $\bar{s}^{(4)}$             & \multicolumn{1}{l|}{$p^{(5)}$} & $\bar{s}^{(5)}$             & \multicolumn{1}{l|}{$p^{(6)}$} & $\bar{s}^{(6)}$             & \multicolumn{1}{l|}{$p^{(7)}$} & $\bar{s}^{(7)}$             \\ \hline
\multicolumn{1}{|c|}{227}       & \multicolumn{1}{c|}{0.0944} & \multicolumn{1}{c|}{1580}      & \multicolumn{1}{c|}{0.0108} & \multicolumn{1}{c|}{6831}      & \multicolumn{1}{c|}{0.0022} & \multicolumn{1}{c|}{11949}     & \multicolumn{1}{c|}{0.0012} & \multicolumn{1}{c|}{16757}     & \multicolumn{1}{c|}{0.0008} & \multicolumn{1}{c|}{19249}     & \multicolumn{1}{c|}{0.0007} \\ \hline
\end{tabular}\label{table:character}
\end{table}

\subsection{HCC and DRG code groupings}
%Hierarchical Condition Categories (HCC) is an established way of grouping of ICD-10 codes, and it has been employed in practical risk adjustment models. The many-to-one mappings between ICD-10 CM codes and HCC codes can be found in the CMS website~\cite{HCC}. In the FY2018 mapping, there are 75 unique HCC codes while only 4015 ICD-10 CM codes are included. For instance, HCC 15 contains three codes: C9210, C9211, and C9212. In order to compare the model predictability as well as the coefficient consistency using different ways of code grouping, we need to extract the design matrix $X_{HCC}$ from the original $X$ using the mentioned crosswalk. Panel A in Fig.~\ref{HCCDRG2018} displays the distribution over the frequencies of the HCC code. Comparing with Panel F in Fig.~\ref{HessianDigHistVSLetters2018}, it can be seen that the power law behavior, i.e. number of less frequent ICD-10 codes is exponentially more than those of more frequent codes, is suppressed by the HCC scheme.

\begin{figure}[ht]
  \centering
  \includegraphics[width=0.8\textwidth]{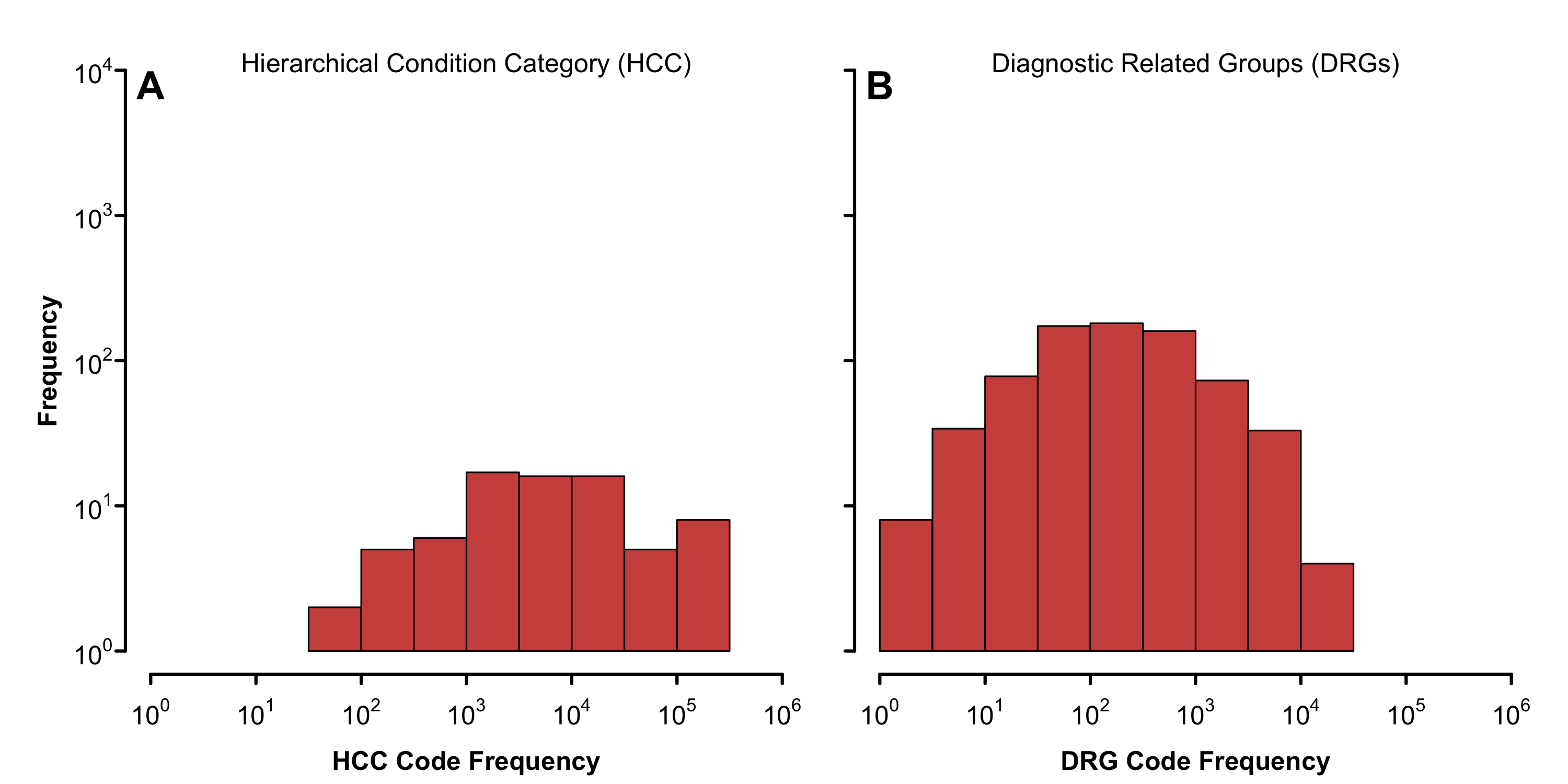}
  \caption{Left panel: Histogram of HCC codes frequencies. Right panel: Histogram of DRG codes frequencies. Code groupings reduce the prevalence of rare codes in the ICD-10 representation of diagnoses.}
  \label{HCCDRG2018}
\end{figure}

Hierarchical Condition Categories (HCC) offers a standardized method for grouping ICD-10 codes and is widely used in risk adjustment models. 
A many-to-one mapping exists between ICD-10-CM codes and HCC codes, as documented by the Centers for Medicare \& Medicaid Services (CMS)~\cite{HCC}. 
There are 75 unique HCC codes, but only 4015 ICD-10-CM codes are included in these HCCs. 
The others are discarded.
The left panel in Fig.~\ref{HCCDRG2018} illustrates the frequency distribution of HCC codes appeared in our subset. In contrast to Panel F of Fig.~\ref{HessianDigHistVSLetters2018}, the HCC-based representation attenuates the power-law distribution observed in ICD-10 frequencies, where rare codes vastly outnumber common ones.

%Another way of grouping and discarding diagnostic codes is Diagnosis-Related Groups (DRG). Such scheme maps the principal diagnostic codes to one DRG code. In MedPAR data, this information is available. The total number of unique DRG codes extracted from the MedPAR subset is 744. Panel B in Fig.~\ref{HCCDRG2018} shows the distribution of the DRG frequencies. Similarly, the abundance of rare codes in ICD-10 representation is not seen here.

An alternative approach to aggregating diagnostic information is through Diagnosis-Related Groups (DRGs), which map principal diagnosis codes to a single DRG code. 
The DRG code for inpatient stay is available within the MedPAR dataset. 
From the subset analyzed, we identify 744 unique DRG codes. 
The right panel of Fig.~\ref{HCCDRG2018} depicts the frequency distribution of these DRG codes. 
Similarly to the HCC codes, the long-tail pattern characteristic of ICD-10 code frequencies is notably diminished, indicating a reduction in the prevalence of rare codes under the DRG scheme.

%have additionally considered two models of log stay costs against the accompanying HCC codes and DRG codes, respectively.

\section{Results}\label{results} 

This section presents the results of our regression analyses using the FY2018 dataset. 
We evaluate the log-linear model's predictive performance and consistency across varying levels of code granularity and regularization strengths. 
As comparisons, the results from using the HCC and DRG code grouping schemes are also presented. In addition, the accuracies from employing decision tree and random forest are provided.
Our analyses utilize the corresponding implementations through the Scikit-Learn library~\cite{scikit-learn}.

\subsection{Model Accuracy and Effective Dimension}

\begin{figure}[ht]
  \centering
  \includegraphics[width=0.9\textwidth]{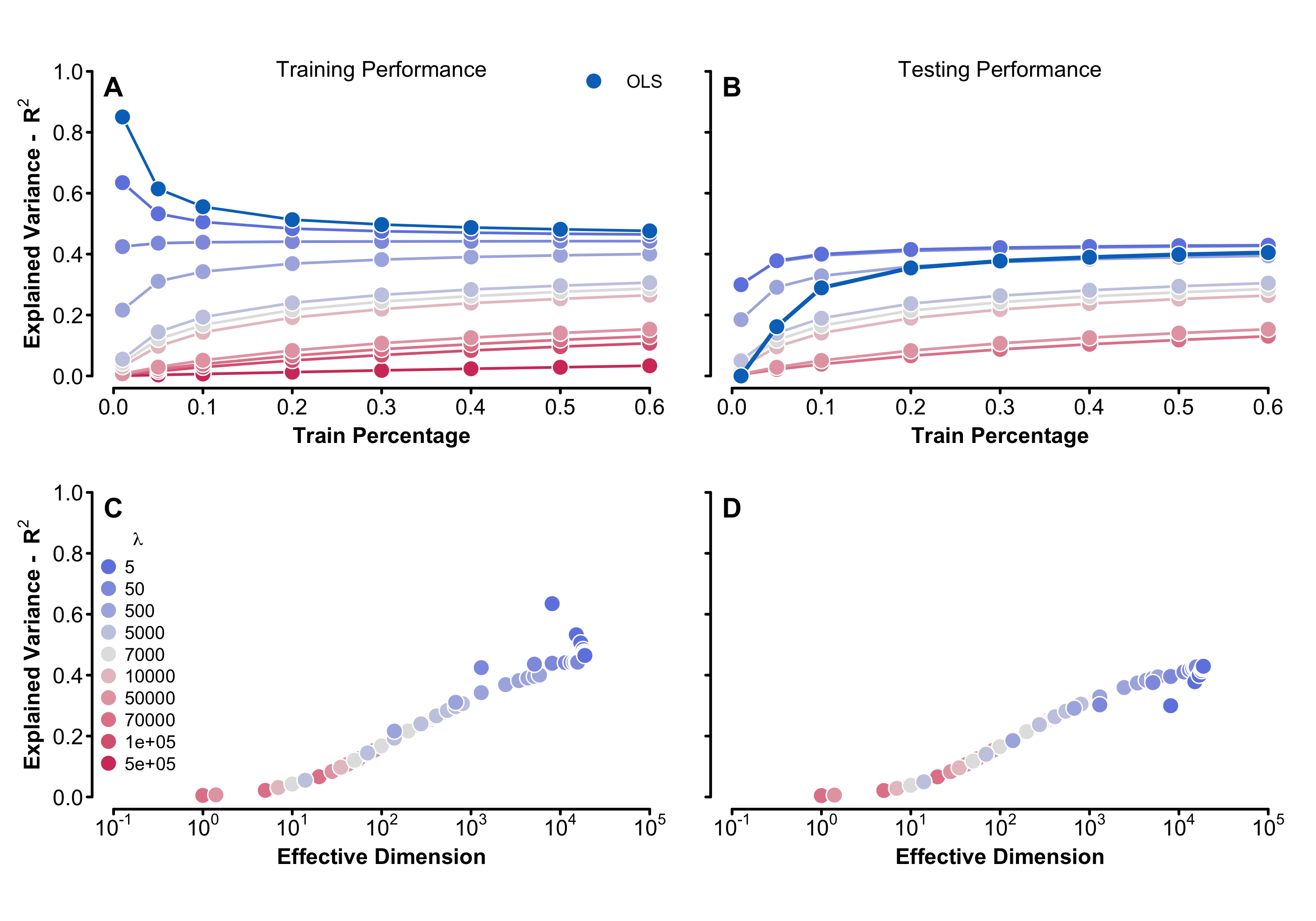}
\caption{Training ({\bf A}) and test ({\bf B}) scores for Ridge models with varying training size ratio and regularization $\lambda$. Panels ({\bf C}) and ({\bf D}) display these training and test, respectively, scores against the effective dimension upper bound $\rho_B$.}
  \label{AccVSEffDPercenatge}
\end{figure}

With the extracted design matrices $X^{(l)}$, it is straightforward to apply LinearRegression (OLS) and Ridge to estimate model coefficients and compute training and test $R^2$ scores. 
We vary the training set size by sampling training ratios [${0.01, 0.05, 0.1, 0.2, 0.3, 0.4, 0.5, 0.6}$] from the full FY2018 dataset of $n = 495,023$ hospital stays. 
We first focus on the highest granularity case, $l = 7$.
Figure \ref{AccVSEffDPercenatge} summarizes the results for Ridge models trained with regularization strengths $\lambda = \{5, 50, 500, 5\text{E}3, 7\text{E}3, 1\text{E}4, 5\text{E}4, 7\text{E}4, 1\text{E}5, 5\text{E}5\}$. 
Training scores are shown in panel A, with OLS results in brightest blue for reference, while panel B shows the corresponding test scores. 
Overfitting is observed only for the smallest training size (approximately 5,000) when $n_{tr}<p$ and regularization is weakest -- evidenced by a large discrepancy between training and test scores. 
As expected, increasing regularization suppresses both scores, as the estimated $\hat{\beta}$ is increasingly biased toward zero \cite{mel2021theory}.
Notably, both training and test scores decrease with smaller training ratios. 
This behavior arises because Scikit-Learn’s Ridge implementation minimizes the unscaled objective, $\mathcal{L} = \sum (y_i - \hat{y}_i)^2 + \lambda \|\beta\|^2$, making the regularization term effectively stronger as the training set shrinks. 
However, when scores are plotted against the effective dimension upper bound $\rho_B = \frac{(p+1) \bar{s}}{\bar{s} + \lambda / n_{\text{tr}}}$, the results largely collapse onto a single curve, as shown in panels C and D of Figure \ref{AccVSEffDPercenatge}.
Deviations from this collapse suggest that the sample covariance $X^\prime X/n$ may be a noisy estimator of the true covariance for small training sizes.

\begin{figure}[ht]
  \centering
  \includegraphics[width=1\textwidth]{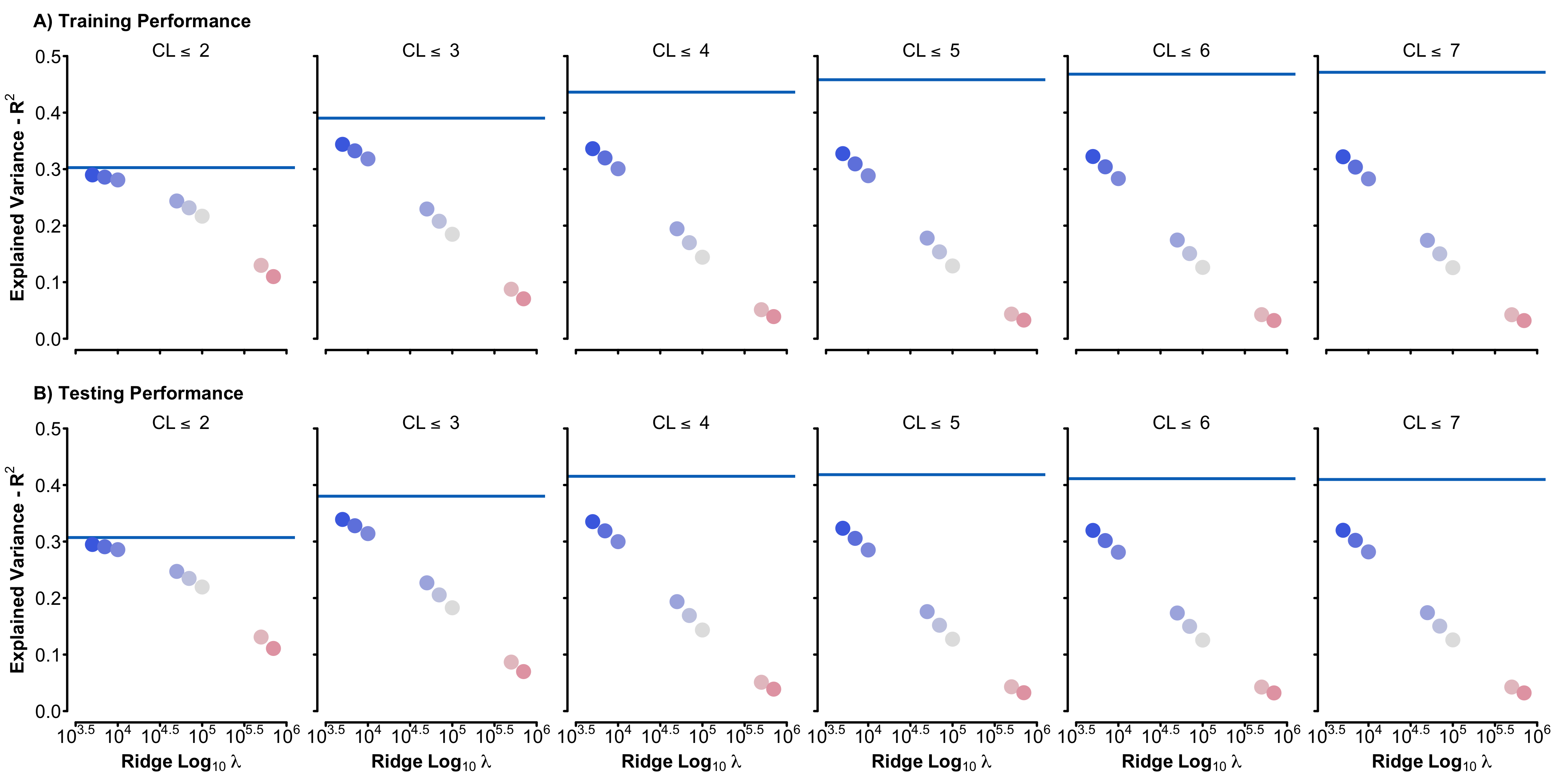}
    \caption{Training (top) and test (bottom) $R^2$ scores for log-linear models with different granularity levels specified by $CL\leq l=[2,3,4,5,6,7]$.}
  \label{CLVaryingAcc}
\end{figure}

Next, we examine how prediction performance varies with code granularity. 
Figure \ref{CLVaryingAcc} displays training and test $R^2$ scores for Ridge models fit to $X^{(l)}$ for $l = 2$ to 7, with a fixed training ratio of 0.8. 
Ridge models use regularization strengths $\lambda = \{0.8, 1, 10, 50, 100, 500, 1000\}$. 
Across all granularities, OLS test scores are comparable to those of Ridge, with OLS occasionally outperforming the regularized models at higher $\lambda$. 
Lower granularity leads to smaller design matrices (i.e., fewer columns $p^{(l)}$), which in turn affects both model capacity and regularization sensitivity.
As for the model using HCC code grouping, the training and test $R^2$ scores are both around the level of 0.075. The DRG-based model, on the other hand, has a training score of 0.41 along with a test score around 0.40, which is close to the performance of OLS using the highest granular codes.

%\textbf{I removed the text on trees in this section because it would require you to write a whole section on tree-based models and how they were implemented, which I did not see. (I disagreed. In the literature review, RF models were used in the Thailand paper, and we cited the accuracy. In review reports, RFs were mentioned in almost every reviewer's report. If the goal is to maximized the chance of acceptance, the accuracy must be included. I do not see why we have to have a section devoted to the Method of tree-based mode.)}
%Here, it may be interesting to mention the performance of tree-based models. With highest granularity $l=7$, the regularized decision tree model with min\_sample\_split=100 results in the training and test scores of 0.38 and 0.09, respectively. The Random Forest models with n\_estimators=20 and same regularization yields the scores of 0.46 and 0.29 for training and test, respectively. It is noted that the training OLS, decision tree, and random forest models on sparse data with shape $(n,p)\approx(400K,20K)$ takes 50, 10, and 132 seconds with our desktop.

\subsection{Coefficient Consistency Across Samples}

\begin{figure}[ht]
  \centering
  \includegraphics[width=0.8\textwidth]{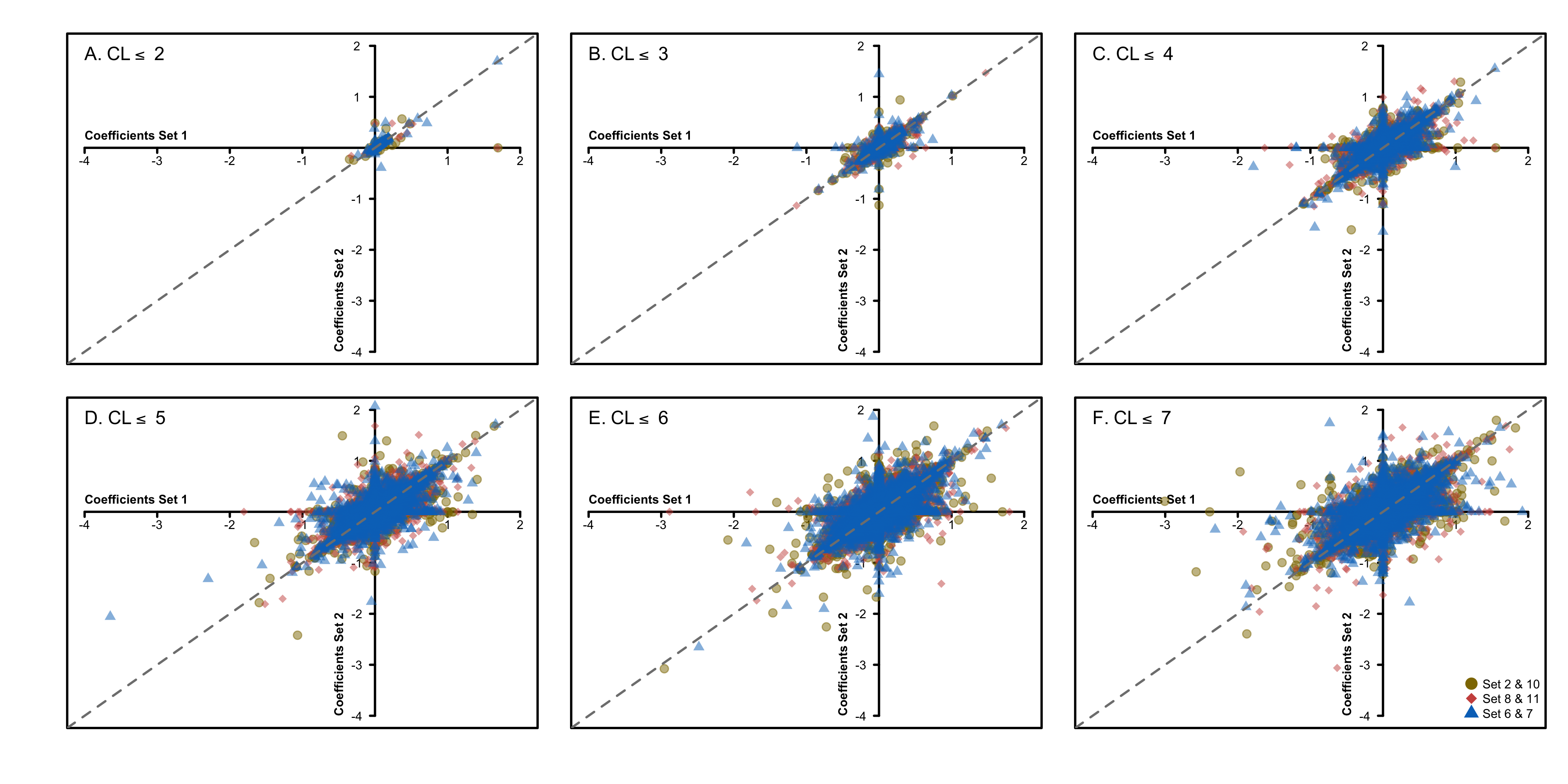}
  \caption{Scatter plotting of one set of regression coefficients against another for measuring the consistency $\eta$ defined in Eq.~\ref{consistency_metric}. Red and blue clouds represent two different pairs. From the top-left panel to right and from bottom-left to right indicate different $CL\leq l=[2,3,4,5,6,7]$.}
  \label{CLVaryingCon}
\end{figure}

\begin{figure}[ht]
  \centering
   \includegraphics[width=0.8\textwidth]{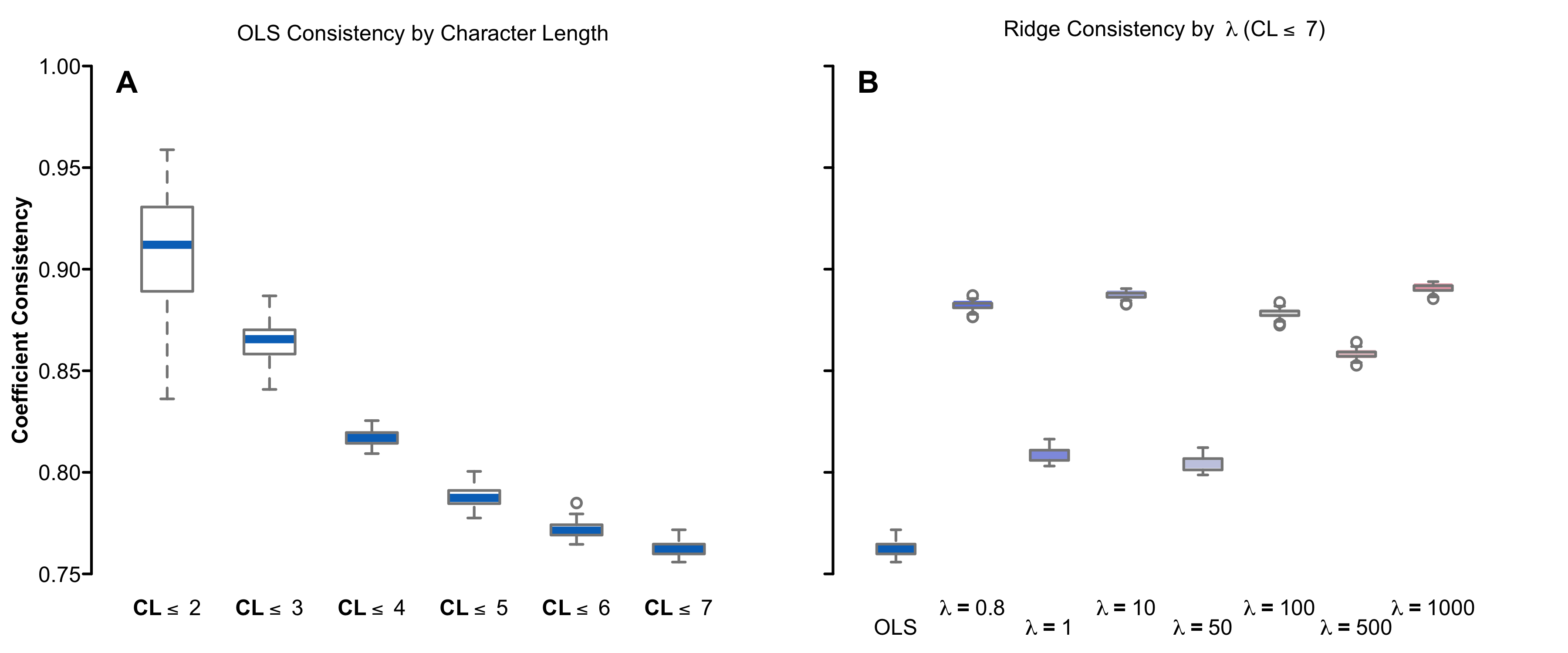}
  \caption{Panel A: OLS coefficient consistency $\eta$ for varying granularity $CL\leq[2,3,4,5,6,7]$ with training ratio 0.8. B: the consistency for varying regularization $\lambda$ in Ridge with code granularity $CL\leq 7$.}
  \label{Consistency2018}
\end{figure}

We assess the consistency of coefficient estimates across different training subsets by computing the consistency metric $\eta$ (Eq. \ref{consistency_metric}) using an ensemble of 10 OLS models fit to randomly drawn training sets from the FY2018 data.
A consistent model should yield similar $\hat{\beta}$ vectors across different samples. 
This can be visualized with a scatterplot where one coefficient vector is drawn against the other.
Perfect agreement corresponds to all points falling on the diagonal.
Figure \ref{CLVaryingCon} shows scatter plots for two sample pairs of coefficient vectors across all granularity levels. 
At the finest granularity ($l = 7$, bottom right), the coefficient clouds form rotated ellipses aligned with the diagonal, indicating {\it weak} agreement between samples. 
As granularity decreases (moving from bottom right to top left), the number of predictors $p^{(l)}$ drops, and the coefficient magnitudes shrink, mimicking the behavior of explicit regularization despite no penalty being applied in OLS.
Importantly, the coefficient clouds are better explained by the diagonal line at lower granularity, meaning there is \textit{strong} agreement between samples.
To quantify this behavior, we compute $\eta$ using the Spearman correlation from the SciPy library \cite{2020SciPy-NMeth}. 
Figure \ref{Consistency2018} displays the results. 
The left panel shows an increasing trend of $\eta$ as the granularity is reduced in OLS models, suggesting more stable coefficients in lower-dimensional settings.
The right panel displays consistency for Ridge models at $l = 7$, across eight regularization strengths $\lambda / n_{\text{tr}}=\{0, 2\text{E}{-6}, 2.5\text{E}{-6}, 
2.5\text{E}{-5}, 1.3\text{E}{-4}, 2.5\text{E}{-4}, 1.3\text{E}{-3}, 2.5\text{E}{-3}\}$.
Even small regularization leads to a noticeable jump (0.05) in $\eta$ over OLS. As $\lambda$ increases, consistency improves, reaching values near 0.9 for the strongest penalties.
%This gain in stability arises from the sparsity of the Hessian $X^\prime X$. %wrong
Since many ICD codes are rare, the diagonal entries of the Hessian—approximating eigenvalues—vary significantly. 
In the Ridge solution, $\left(\tilde{X}^\prime \tilde{X} / n_{\text{tr}} + \lambda / n_{\text{tr}} \right)^{-1}$, the penalty term has minimal effect on coefficients associated with frequent codes (e.g., appearing 10,000+ times), but strongly dampens coefficients for infrequent ones. 
This selective shrinkage reduces coefficient variance and improves inter-sample agreement, hence increasing $\eta$.

\begin{figure}[ht]
  \centering
  \includegraphics[width=0.8\textwidth]{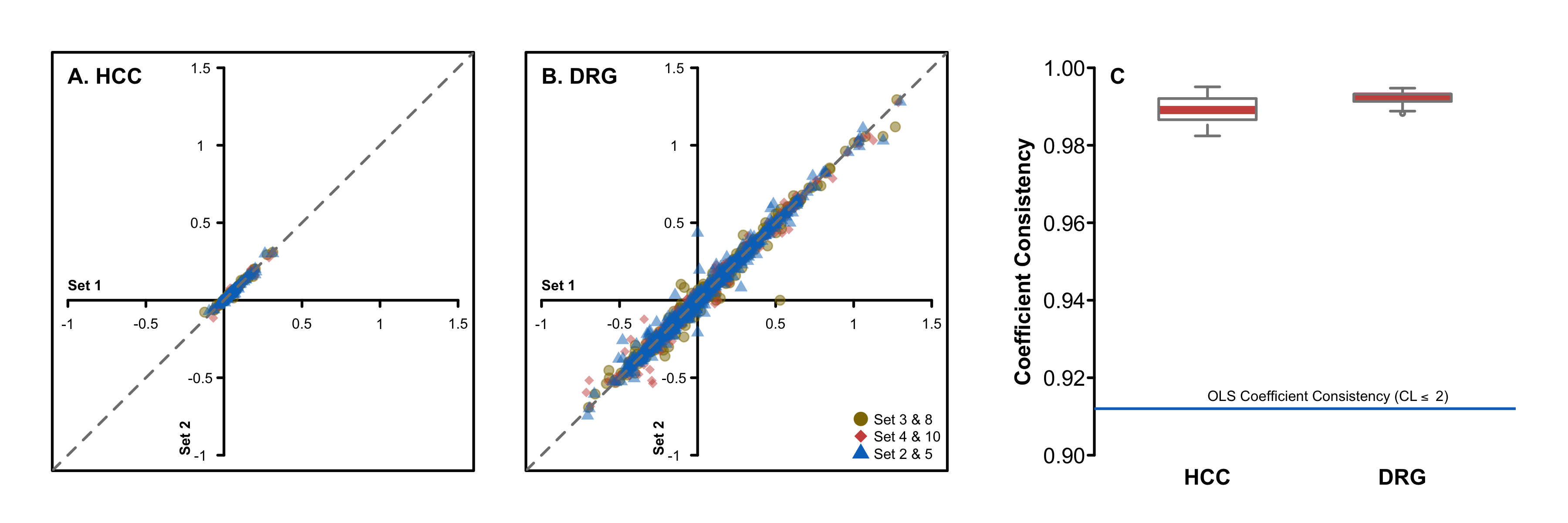}
  \caption{Left panel: similar to Fig.~\ref{CLVaryingCon}, one set of the regression coefficients using HCC code groupings is scatter-plotted against another. Quantitative agreement between coefficients from training OLS with different subsamples is shown if the point clouds are concentrated to the diagonal line. Middle panel: result from employing DRG scheme. Right panel: the coefficient consistency $\eta$ for both established schemes.}
  \label{ConsistencyHCCDRG}
\end{figure}

As a comparison, the HCC and DRG code grouping schemes are also employed to fit the log cost with the method of OLS. The number of predictors is reduced from 19249 (the finest granularity level using ICD-10 code) to 75 (HCC) and 744 (DRG), respectively. We shall emphasize that the design matrix $X_{HCC}$ and $X_{DRG}$ remain binary while the $X$'s for lower granularity levels ($l<7$) can have entries larger than unity in the process of variable aggregation. Unlike HCC and DRG, our scheme of code grouping does not discard any ICD code. The left panel in Fig.~\ref{ConsistencyHCCDRG} displays the quantitative agreement between regression coefficients from training with different subsamples. The middle panel in Fig.~\ref{ConsistencyHCCDRG} shows the result for DRG grouping. As both established schemes suppress the abundance of rare codes, the obtained consistency $\eta$ in the right panel demonstrates improvement from directly employing the ICD-10 codes as predictors.  
%The predictive accuracies are not superior to the OLS using the most granular ICD-10 codes. 

\section{Discussion}\label{discussions} 

This study investigated how varying ICD code granularity and model regularization affected prediction accuracy and coefficient stability in log-linear models fitted to healthcare (inpatient) data. 
Our central finding is that reducing code granularity implicitly regularizes the model, even without explicit penalty terms. 
As granularity decreases, the number of predictors $p^{(l)}$ drops, resulting in coefficient shrinkage and improved stability, similar to the effects of Ridge regression.
We quantified this effect using a coefficient consistency metric $\eta$, which increases with explicit regularization strength and decreasing granularity.
Our finding demonstrates that code aggregation is a powerful, interpretable form of implicit regularization, which offers a practical strategy when data are high-dimensional and sparse. 
This is particularly relevant in healthcare settings, where maintaining model interpretability and robustness is essential.

%In terms of the accuracy, the OLS models have highest test accuracy 0.41 when the code has the highest granularity. The accuracy gradually decreases to around 0.3 when the code granularity is $CL\leq2$. OLS along with HCC grouping schemes results in much worst test accuracy than OLS with highest granularity ICD-10 codes. Surprisingly, OLS with DRG predictors has similar test accuracy as OLS with ICD-10 codes. As for consistency $\eta$, OLS with decreasing code granularity in ICD-10 codes does show improvement from around 0.75 to 0.9. Ridge models and clinical grouping schemes all show superior performance in $\eta$. 

With respect to predictive accuracy, the OLS models achieve their highest test accuracy (0.41) when using ICD-10 codes at the finest level of granularity. Accuracy declines gradually, reaching approximately 0.3 when the code granularity is reduced to $CL \leq 2$. Combining OLS with HCC grouping schemes yields even poorer accuracy compared to OLS with the most granular ICD-10 codes. Interestingly, OLS models with DRG predictors achieve accuracy levels comparable to those with ICD-10 codes. In terms of consistency ($\eta$), however, OLS models benefit from reduced ICD-10 granularity, improving from roughly 0.75 to 0.9. Ridge models and clinical grouping schemes, by contrast, consistently outperform OLS in $\eta$.

Our analysis of the Hessian matrix $X^{(7){\prime}}X^{(7)}$ revealed that its diagonal entries follow a power-law distribution ($\alpha \approx 1.93$), reflecting a highly skewed code frequency landscape. 
This structure leads to instability in estimates for low-frequency codes, which can be mitigated by aggregating codes or applying explicit regularization.
We also introduced the effective dimension upper bound $\rho_B$, which unifies model accuracy across varying sample sizes and regularization levels. 
Plotting $R^2$ scores against $\rho_B$ led to an empirical collapse of performance curves, suggesting it as a meaningful proxy for model capacity in sparse settings.

%Toward a more practical implementation of code grouping for the purpose of reducing $\eta$, several issues remain open for future research. As accuracy and consistency are both critical for implementing machine learning models in healthcare domain, further research aiming to balance the two factors is relevant. Obviously, this work does not include demographic variables such as sex and age because the goal is to understand the implication of high dimensional sparse binary variables to the consistency of regression coefficients. Subgroup analysis regarding model accuracy and consistency is particular important to application in healthcare system. As for theoretical problems, it is interesting to model the distribution from which the highly correlated binary ICD-10 codes for inpatients are sampled, and computing the bound for test accuracy with such distribution is desirable.  

For a more practical implementation of code grouping aimed at reducing $\eta$, several important questions remain for future investigation. Given the critical importance of both accuracy and consistency in deploying machine learning models within the healthcare domain, future work should focus on strategies to balance these two aspects effectively. This study intentionally omits demographic variables in the section of Methods, as its primary goal is to examine how high-dimensional, sparse binary variables influence the consistency of regression coefficients. However, conducting subgroup analyses, e.g. races and specialities, to evaluate model accuracy and consistency remains particularly relevant for real-world healthcare applications. On the theoretical front, a promising direction involves modeling the underlying distribution from which highly correlated binary ICD-10 codes for inpatients are drawn. Deriving accuracy bounds~\cite{liao2020random,mel2021theory,hastie2022surprises} for models trained on such distributions would also be a valuable contribution.

\paragraph{Limitations}

One important limitation of our approach is that merging ICD codes solely based on code granularity (i.e., truncating or aggregating codes by character length) represents just one of many possible strategies for reducing the dimensionality of diagnosis-based models. 
While this method provides a systematic and reproducible framework that naturally introduces implicit regularization, it does not incorporate any clinical or domain-specific knowledge about which codes may be functionally or causally related.
In real-world cost prediction systems, such as the CMS-HCC risk adjustment model, diagnostic codes are grouped using extensive clinical expertise to capture meaningful associations with healthcare utilization and outcomes.
These manually curated groupings aim to ensure medical interpretability and policy relevance, which our purely structural approach does not attempt to replicate.

Moreover, the space of possible groupings is combinatorially large, and choosing among them has substantial implications for model performance, interpretability, and fairness. 
Our work treats granularity as a proxy for grouping, but exploring optimal groupings in a data-driven way while maintaining consistency and accuracy remains an open and important challenge.
Future work could explore automated grouping mechanisms by optimizing for coefficient stability and predictive power across resampled datasets. This is conceptually similar to learned pooling in computer vision, where hierarchical features are grouped to improve both robustness and generalization \cite{sun2017learning}. 
In healthcare, such approaches may lead to more flexible, interpretable, and adaptive models, especially when dealing with noisy or incomplete clinical coding.

\section{Conclusions}\label{conclusion}

%We applied ordinary least square method in modeling log of stay cost with sparse indicators of ICD-10 codes. The problem of inconsistent regression coefficients across training with different subsamples is identified. We proposed a metric for quantifying the inconsistency using Spearman correlation. 
%As the OLS coefficients can quantify the individual patient's health risk, the consistency is a critical issue as well in addition to accuracy and interpretability when applying machine learning models in healthcare.   
%A solution to mitigating the issue is to truncate the ICD-10 codes so that the regression variables are grouped and added. Mathematically, the small eigenvalues of Hessian matrix causes strong variance in estimating the OLS regression coefficients. The grouping by truncating ICD-10 codes gradually increases the trace of Hessian matrices as the code granularity is lowered, which can be viewed as an implicit regularization for improving the coefficient consistency in OLS. Observing the evolving distributions over the diagonal entries of Hessian matrices, the abundance of rare codes is alleviated by reducing the code granularity. Schemes of code groupings such as HCC and DRG also prevent such abundance of rare codes from appearing.   

In this study, we employed the ordinary least squares (OLS) method to model the logarithm of hospital stay costs using sparse indicators derived from ICD-10 codes. We identified a key issue: the instability of regression coefficients across different training subsamples. To address this, we proposed a metric based on Spearman correlation to quantify coefficient inconsistency.
Since OLS coefficients can serve as quantifications of individual patient health risks, ensuring consistency is as crucial as achieving accuracy and interpretability in healthcare-related machine learning models. To mitigate inconsistency, we introduced a truncation approach that groups ICD-10 codes, thereby aggregating regression variables. From a mathematical standpoint, small eigenvalues in the Hessian matrix is due to infrequent codes, contributing to high variance in OLS coefficient estimation. By reducing the granularity of ICD-10 codes through grouping, we effectively increase the trace of the Hessian matrix, acting as an implicit form of regularization that enhances coefficient stability.
Furthermore, by examining the distribution of diagonal entries in the Hessian matrices, we observed that reducing code granularity alleviates the overrepresentation of rare codes. Established grouping schemes such as Hierarchical Condition Categories (HCC) and Diagnosis-Related Groups (DRG) similarly help mitigate the sparsity caused by infrequent codes, contributing to robust, interpretable, and practical risk adjustment models.

\section*{Acknowledgements and Funding Sources} The encouragements and suggestions from anonymous reviewers of MLHC 2025 are greatly appreciated. This work was supported by The Analytics for Equity Initiative under National Science Foundation (award number: 49100423C0041).  
%https://nsf-gov-resources.nsf.gov/2022-06/Summary%20of%20the%20Analytics%20for%20Equity%20Initiative.pdf
%\newpage
% Use the style of numbering in square brackets.
% If nothing is used, default style will be taken.
%\begin{enumerate}[a)]
%\item 
%\item 
%\item 
%\end{enumerate}  

\section*{Declaration of competing interest} The authors declare that they have no known competing financial interests or personal relationships that could have appeared to influence the work reported in this paper.

\section*{Data availability} The data used in this study are from the Medicare Provider Analysis and Review (MedPAR) file, which contains administrative claims data for services provided to Medicare beneficiaries. These data are not publicly available due to patient privacy protections but can be obtained through a data use agreement with the Centers for Medicare \& Medicaid Services (CMS). Interested researchers may request access by submitting a research protocol and data request through the CMS Research Data Assistance Center (ResDAC).
%\begin{itemize}
%\item 
%\item 
%\item 
%\end{itemize}  

%\begin{description}
%\item[]
%\item[] 
%\item[] 
%\end{description}  

%\clearpage %%Remove this from your manuscript

% Figure
%\begin{figure}%[]
%  \centering
%    \includegraphics{}
%    \caption{}\label{fig1}
%\end{figure}

%\begin{table}%[]
%\caption{}\label{tbl1}
%\begin{tabular*}{\tblwidth}{@{}LL@{}}
%\toprule
%  &  \\ % Table header row
%\midrule
% & \\
% & \\
% & \\
% & \\
%\bottomrule
%\end{tabular*}
%\end{table}

% Uncomment and use as the case may be
%\begin{theorem} 
%\end{theorem}

% Uncomment and use as the case may be
%\begin{lemma} 
%\end{lemma}


%\section{}\label{}

%\newpage
%\bibliographystyle{cas-model2-names}

%\bibliography{example_paper.bib}

\begin{thebibliography}{10}

\bibitem{Duan01041983}
Naihua Duan, Willard~G. Manning, Carl~N. Morris, and Joseph P.~Newhouse and.
\newblock A comparison of alternative models for the demand for medical care.
\newblock {\em Journal of Business \& Economic Statistics}, 1(2):115--126,
  1983.

\bibitem{griswold2004analyzing}
Michael Griswold, Giovanni Parmigiani, Arnie Potosky, and Joseph Lipscomb.
\newblock Analyzing health care costs: a comparison of statistical methods
  motivated by medicare colorectal cancer charges.
\newblock {\em Biostatistics}, 1(1):1--23, 2004.

\bibitem{Layton2017}
Timothy~J. Layton.
\newblock Imperfect risk adjustment, risk preferences, and sorting in
  competitive health insurance markets.
\newblock {\em Journal of Health Economics}, 56:259--280, 2017.

\bibitem{fernandez2019estimating}
Iv{\'a}n~S{\'a}nchez Fern{\'a}ndez, Marta Amengual-Gual, Cristina~Barcia
  Aguilar, and Tobias Loddenkemper.
\newblock Estimating the cost of status epilepticus admissions in the {United
  States of America} using {ICD}-10 codes.
\newblock {\em Seizure}, 71:295--303, 2019.

\bibitem{James2013ISLR}
Gareth James, Daniela Witten, Trevor Hastie, and Robert Tibshirani.
\newblock {\em An Introduction to Statistical Learning}.
\newblock Springer Texts in Statistics. Springer New York, NY, July 2021.

\bibitem{kan2019exploring}
Hong~J Kan, Hadi Kharrazi, Hsien-Yen Chang, Dave Bodycombe, Klaus Lemke, and
  Jonathan~P Weiner.
\newblock Exploring the use of machine learning for risk adjustment: A
  comparison of standard and penalized linear regression models in predicting
  health care costs in older adults.
\newblock {\em PloS one}, 14(3):e0213258, 2019.

\bibitem{irvin2020incorporating}
Jeremy~A Irvin, Andrew~A Kondrich, Michael Ko, Pranav Rajpurkar, Behzad
  Haghgoo, Bruce~E Landon, Robert~L Phillips, Stephen Petterson, Andrew~Y Ng,
  and Sanjay Basu.
\newblock Incorporating machine learning and social determinants of health
  indicators into prospective risk adjustment for health plan payments.
\newblock {\em BMC Public Health}, 20:1--10, 2020.

\bibitem{Reed:2001aa}
S~D Reed, D~K Blough, K~Meyer, and J~G Jarvik.
\newblock Inpatient costs, length of stay, and mortality for cerebrovascular
  events in community hospitals.
\newblock {\em Neurology}, 57(2):305--314, Jul 2001.

\bibitem{THONGPETH2021100769}
Wichayaporn Thongpeth, Apiradee Lim, Akemat Wongpairin, Thaworn Thongpeth, and
  Santhana Chaimontree.
\newblock Comparison of linear, penalized linear and machine learning models
  predicting hospital visit costs from chronic disease in thailand.
\newblock {\em Informatics in Medicine Unlocked}, 26:100769, 2021.

\bibitem{sandra2023prediction}
J~Ruth Sandra, Sanjana Joshi, Aditi Ravi, Ashwini Kodipalli, Trupthi Rao, and
  Shoaib Kamal.
\newblock Prediction of cost for medical care insurance by using regression
  models.
\newblock In {\em International Conference on Emerging Research in Computing,
  Information, Communication and Applications}, pages 311--323. Springer, 2023.

\bibitem{RAO2024100351}
A.~Ravishankar Rao, Raunak Jain, Mrityunjai Singh, and Rahul Garg.
\newblock Predictive interpretable analytics models for forecasting healthcare
  costs using open healthcare data.
\newblock {\em Healthcare Analytics}, 6:100351, 2024.

\bibitem{Evans2011Evalutaion}
Gregory~C. Pope, John Kautter, Melvin~J. Ingber, Sara Freeman, Rishi Sekar,
  Cordon Newhart, and Melissa~A. Evans.
\newblock Evaluation of the {CMS-HCC} risk adjustment model.
\newblock Technical report, The Centers for Medicare and Medicaid Services
  Office of Research, Development, and Information, March 2011.

\bibitem{kautter2014hhs}
John Kautter, Gregory~C Pope, Melvin Ingber, Sara Freeman, Lindsey Patterson,
  Michael Cohen, and Patricia Keenan.
\newblock The {HHS-HCC} risk adjustment model for individual and small group
  markets under the affordable care act.
\newblock {\em Medicare \& Medicaid research review}, 4(3):mmrr2014--004, 2014.

\bibitem{Ash2017}
Arlene~S. Ash, Eric~O. Mick, Randall~P. Ellis, Catarina~I. Kiefe, Jeroan~J.
  Allison, and Melissa~A. Clark.
\newblock Social determinants of health in managed care payment formulas.
\newblock {\em JAMA Internal Medicine}, 177(10):1424--1430, 10 2017.

\bibitem{hoerl1970ridge}
Arthur~E. Hoerl and Robert~W. Kennard.
\newblock Ridge regression: Biased estimation for nonorthogonal problems.
\newblock {\em Technometrics}, 12(1):55--67, 1970.

\bibitem{hastie2017elements}
Trevor Hastie, Robert Tibshirani, and Jerome Friedman.
\newblock The elements of statistical learning: data mining, inference, and
  prediction, 2017.

\bibitem{Goodfellow-et-al-2016}
Ian Goodfellow, Yoshua Bengio, and Aaron Courville.
\newblock {\em Deep Learning}.
\newblock MIT Press, 2016.
\newblock \url{http://www.deeplearningbook.org}.

\bibitem{Pfuntner2006}
A~Pfuntner, L~Wier, and C~Steiner.
\newblock {\em Costs for Hospital Stays in the United States}, pages 1--11.
\newblock Number 146. Agency for Healthcare Research and Quality, Rockville
  (MD; US), 2 2006.

\bibitem{CMS}
{Centers for Medicare \& Medicaid Services}.
\newblock {ICD-10 Files \& News Archive}.
\newblock
  \url{https://www.cms.gov/medicare/coding-billing/icd-10-codes/icd-10-cm-icd-10-pcs-gem-archive},
  2025.
\newblock Accessed: July 23, 2025.

\bibitem{Omalley2005}
Kimberly~J. O'Malley, Karon~F. Cook, Matt~D. Price, Kimberly~Raiford Wildes,
  John~F. Hurdle, and Carol~M. Ashton.
\newblock Measuring diagnoses: {ICD} code accuracy.
\newblock {\em Health Services Research}, 40(5p2):1620--1639, 2005.

\bibitem{dicker2014variance}
Lee~H Dicker.
\newblock Variance estimation in high-dimensional linear models.
\newblock {\em Biometrika}, pages 269--284, 2014.

\bibitem{Shiban2021}
Nisreen Shiban, Joshua Gaul, Henry Zhan, Andrew Elhabr, Nima Kokabi,
  Jamlik-Omari Johnson, Tarek Hanna, Justin Schrager, Judy Gichoya, Imon
  Banerjee, and Hari Trivedi.
\newblock Machine learning methods to predict survival in patients following
  traumatic aortic injury.
\newblock {\em medRxiv}, 2021.

\bibitem{mirtchouk2021hierarchical}
Mark Mirtchouk, Bharat Srikishan, and Samantha Kleinberg.
\newblock Hierarchical information criterion for variable abstraction.
\newblock In {\em Machine Learning for Healthcare Conference}, pages 440--460.
  PMLR, 2021.

\bibitem{qiao2022}
Edmund~M. Qiao, Alexander~S. Qian, Vinit Nalawade, Rohith~S. Voora, Nikhil~V.
  Kotha, Lucas~K. Vitzthum, and James~D. Murphy.
\newblock Evaluating high-dimensional machine learning models to predict
  hospital mortality among older patients with cancer.
\newblock {\em JCO Clinical Cancer Informatics}, (6):e2100186, 2022.
\newblock PMID: 35671416.

\bibitem{technologies10060122}
Dimitrios Zikos and Nailya DeLellis.
\newblock Comparison of the predictive performance of medical coding diagnosis
  classification systems.
\newblock {\em Technologies}, 10(6), 2022.

\bibitem{Fetter1986}
Robert~B. Fetter and Jean~L. Freeman.
\newblock Diagnosis related groups: Product line management within hospitals.
\newblock {\em Academy of Management Review}, 11(1):41--54, 1986.

\bibitem{Lynk2001}
William~J. Lynk.
\newblock One {DRG}, one price? the effect of patient condition on price
  variation within drgs and across hospitals.
\newblock {\em International Journal of Health Care Finance and Economics},
  1(2):111--137, 2001.

\bibitem{wagner2016}
Todd~H. Wagner, Anjali Upadhyay, Elizabeth Cowgill, Theodore Stefos, Eileen
  Moran, Steven~M. Asch, and Peter Almenoff.
\newblock Risk adjustment tools for learning health systems: A comparison of
  dxcg and cms-hcc v21.
\newblock {\em Health Services Research}, 51(5):2002--2019, 2016.

\bibitem{Kim2023}
Juyoung Kim, Minsu Ock, In-Hwan Oh, Min-Woo Jo, Yoon Kim, Moo-Song Lee, and
  Sang-il Lee.
\newblock Comparison of diagnosis-based risk adjustment methods for
  episode-based costs to apply in efficiency measurement.
\newblock {\em BMC Health Services Research}, 23(1):1334, 2023.

\bibitem{MANNING2001461}
Willard~G Manning and John Mullahy.
\newblock Estimating log models: to transform or not to transform?
\newblock {\em Journal of Health Economics}, 20(4):461--494, 2001.

\bibitem{Tibshirani1996}
Robert Tibshirani.
\newblock Regression shrinkage and selection via the lasso.
\newblock {\em Journal of the Royal Statistical Society: Series B
  (Methodological)}, 58(1):267--288, 12 2018.

\bibitem{bakin1999adaptive}
Sergey Bakin.
\newblock {\em Adaptive regression and model selection in data mining
  problems}.
\newblock PhD thesis, The Australian National University, 1999.

\bibitem{yuan2006model}
Ming Yuan and Yi~Lin.
\newblock Model selection and estimation in regression with grouped variables.
\newblock {\em Journal of the Royal Statistical Society: Series B (Statistical
  Methodology)}, 68(1):49--67, 2006.

\bibitem{Taloba2022}
Ahmed~I. Taloba, Rasha~M. Abd El-Aziz, Huda~M. Alshanbari, and Abdal-Aziz~H.
  El-Bagoury.
\newblock Estimation and prediction of hospitalization and medical care costs
  using regression in machine learning.
\newblock {\em Journal of Healthcare Engineering}, 2022(1):7969220, 2022.

\bibitem{Langenberger2023}
Benedikt Langenberger, Timo Schulte, and Oliver Groene.
\newblock The application of machine learning to predict high-cost patients: A
  performance-comparison of different models using healthcare claims data.
\newblock {\em PLOS ONE}, 18(1):1--16, 01 2023.

\bibitem{yang2018}
Chengliang Yang, Chris Delcher, Elizabeth Shenkman, and Sanjay Ranka.
\newblock Machine learning approaches for predicting high cost high need
  patient expenditures in health care.
\newblock {\em BioMedical Engineering OnLine}, 17(1):131, 2018.

\bibitem{RAHMAN2025100411}
Yead Rahman and Prerna Dua.
\newblock A machine learning framework for predicting healthcare utilization
  and risk factors.
\newblock {\em Healthcare Analytics}, 8:100411, 2025.

\bibitem{mel2021theory}
Gabriel Mel and Surya Ganguli.
\newblock A theory of high dimensional regression with arbitrary correlations
  between input features and target functions: sample complexity, multiple
  descent curves and a hierarchy of phase transitions.
\newblock In {\em International Conference on Machine Learning}, pages
  7578--7587. PMLR, 2021.

\bibitem{hastie2022surprises}
Trevor Hastie, Andrea Montanari, Saharon Rosset, and Ryan~J Tibshirani.
\newblock Surprises in high-dimensional ridgeless least squares interpolation.
\newblock {\em Annals of statistics}, 50(2):949, 2022.

\bibitem{maddox2020rethinking}
Wesley~J Maddox, Gregory Benton, and Andrew~Gordon Wilson.
\newblock Rethinking parameter counting in deep models: Effective
  dimensionality revisited.
\newblock {\em arXiv preprint arXiv:2003.02139}, 2020.

\bibitem{aich2018global}
Shubhra Aich and Ian Stavness.
\newblock Global sum pooling: A generalization trick for object counting with
  small datasets of large images.
\newblock {\em arXiv preprint arXiv:1805.11123}, 2018.

\bibitem{sun2017learning}
Manli Sun, Zhanjie Song, Xiaoheng Jiang, Jing Pan, and Yanwei Pang.
\newblock Learning pooling for convolutional neural network.
\newblock {\em Neurocomputing}, 224:96--104, 2017.

\bibitem{harris2020array}
Charles~R. Harris, K.~Jarrod Millman, St{\'{e}}fan~J. van~der Walt, Ralf
  Gommers, Pauli Virtanen, David Cournapeau, Eric Wieser, Julian Taylor,
  Sebastian Berg, Nathaniel~J. Smith, Robert Kern, Matti Picus, Stephan Hoyer,
  Marten~H. van Kerkwijk, Matthew Brett, Allan Haldane, Jaime~Fern{\'{a}}ndez
  del R{\'{i}}o, Mark Wiebe, Pearu Peterson, Pierre G{\'{e}}rard-Marchant,
  Kevin Sheppard, Tyler Reddy, Warren Weckesser, Hameer Abbasi, Christoph
  Gohlke, and Travis~E. Oliphant.
\newblock Array programming with {NumPy}.
\newblock {\em Nature}, 585(7825):357--362, September 2020.

\bibitem{clauset2009power}
Aaron Clauset, Cosma~Rohilla Shalizi, and Mark~EJ Newman.
\newblock Power-law distributions in empirical data.
\newblock {\em SIAM review}, 51(4):661--703, 2009.

\bibitem{mahoney2019traditional}
Michael Mahoney and Charles Martin.
\newblock Traditional and heavy tailed self regularization in neural network
  models.
\newblock In {\em International Conference on Machine Learning}, pages
  4284--4293. PMLR, 2019.

\bibitem{HCC}
{Centers for Medicare \& Medicaid Services}.
\newblock {2018 Model Software \/ICD-10 Mappings}.
\newblock
  \url{https://www.cms.gov/medicare/health-plans/medicareadvtgspecratestats/risk-adjustors-items/risk2018},
  2025.
\newblock Accessed: July 23, 2025.

\bibitem{scikit-learn}
F.~Pedregosa, G.~Varoquaux, A.~Gramfort, V.~Michel, B.~Thirion, O.~Grisel,
  M.~Blondel, P.~Prettenhofer, R.~Weiss, V.~Dubourg, J.~Vanderplas, A.~Passos,
  D.~Cournapeau, M.~Brucher, M.~Perrot, and E.~Duchesnay.
\newblock Scikit-learn: Machine learning in {P}ython.
\newblock {\em Journal of Machine Learning Research}, 12:2825--2830, 2011.

\bibitem{2020SciPy-NMeth}
Pauli Virtanen, Ralf Gommers, Travis~E. Oliphant, Matt Haberland, Tyler Reddy,
  David Cournapeau, Evgeni Burovski, Pearu Peterson, Warren Weckesser, Jonathan
  Bright, St{\'e}fan~J. {van der Walt}, Matthew Brett, Joshua Wilson, K.~Jarrod
  Millman, Nikolay Mayorov, Andrew R.~J. Nelson, Eric Jones, Robert Kern, Eric
  Larson, C~J Carey, {\.I}lhan Polat, Yu~Feng, Eric~W. Moore, Jake
  {VanderPlas}, Denis Laxalde, Josef Perktold, Robert Cimrman, Ian Henriksen,
  E.~A. Quintero, Charles~R. Harris, Anne~M. Archibald, Ant{\^o}nio~H. Ribeiro,
  Fabian Pedregosa, Paul {van Mulbregt}, and {SciPy 1.0 Contributors}.
\newblock {{SciPy} 1.0: Fundamental Algorithms for Scientific Computing in
  Python}.
\newblock {\em Nature Methods}, 17:261--272, 2020.

\bibitem{liao2020random}
Zhenyu Liao, Romain Couillet, and Michael~W Mahoney.
\newblock A random matrix analysis of random fourier features: beyond the
  gaussian kernel, a precise phase transition, and the corresponding double
  descent.
\newblock {\em Advances in Neural Information Processing Systems},
  33:13939--13950, 2020.

\end{thebibliography}
%\bibliographystyle{plainnat}
\bibliographystyle{unsrt}
\end{document}